\documentclass[sigconf]{acmart}

\AtBeginDocument{%
  }

\acmConference[KDD '26]{Proceedings of the 32nd ACM SIGKDD Conference on Knowledge Discovery and Data Mining V.2}{August 09--13, 2026}{Jeju Island, Republic of Korea}

\usepackage{amsfonts}
\usepackage{amsthm}     
\usepackage{bm}         

\usepackage{algorithm}
\usepackage{algorithmic}
\usepackage{multirow}

\newcommand{\R}{\mathbb{R}}             
\newcommand{\E}{\mathbb{E}}             

\newcommand{\bA}{\boldsymbol{A}}
\newcommand{\bQ}{\boldsymbol{Q}}
\newcommand{\bU}{\boldsymbol{U}}
\newcommand{\bO}{\boldsymbol{O}}
\newcommand{\bS}{\boldsymbol{S}}
\newcommand{\bH}{\boldsymbol{H}}

\newcommand{\bX}{\boldsymbol{X}}
\newcommand{\bY}{\boldsymbol{Y}}

\newcommand{\bW}{\boldsymbol{W}}
\newcommand{\bx}{\boldsymbol{x}}
\newcommand{\by}{\boldsymbol{y}}

\newcommand{\bb}{\boldsymbol{b}}

\newcommand{\bz}{\boldsymbol{z}}
\newcommand{\btheta}{\boldsymbol{\theta}}

\newcommand{\bphi}{\boldsymbol{\phi}}
\newcommand{\bpsi}{\boldsymbol{\psi}}
\newcommand{\bomega}{\boldsymbol{\omega}}
\newcommand{\bOmega}{\boldsymbol{\Omega}}
\newcommand{\bgamma}{\boldsymbol{\gamma}}

\newcommand{\cD}{\mathcal{D}}  
\newcommand{\cL}{\mathcal{L}}  
\newcommand{\cM}{\mathcal{M}}  
\newcommand{\cE}{\mathcal{E}}  
\newcommand{\cP}{\mathcal{P}}  

\newcommand{\ie}{\textit{i}.\textit{e}., }
\newcommand{\eg}{\textit{e}.\textit{g}.\ }
\usepackage[capitalize,noabbrev]{cleveref}

\begin{document}

\title{iLTM: Integrated Large Tabular Model}

\author{%
  David Bonet$^{1,2,*}$
  \quad
  Marçal Comajoan Cara$^{1,3,*}$
  \quad
  Alvaro Calafell$^{1}$
  \\
  Daniel Mas Montserrat$^{1,2}$
  \quad
  Alexander G. Ioannidis$^{1,2}$
  \\\vspace{4mm}\normalsize
  $^1$Stanford University \quad $^2$University of California, Santa Cruz \\ $^3$University of California, Berkeley\\\vspace{2mm}
  $^*$ Equal contribution.\quad Correspondence: \texttt{ioannidis@stanford.edu}
}

\renewcommand{\shortauthors}{David Bonet, Marçal Comajoan Cara, Alvaro Calafell, Daniel Mas Montserrat, \& Alexander G. Ioannidis}

\begin{abstract}
Tabular data underpins decisions across science, industry, and public services. Despite rapid progress, advances in deep learning have not fully carried over to the tabular domain, where gradient-boosted decision trees (GBDTs) remain a default choice in practice. We present iLTM, an integrated Large Tabular Model that unifies tree-derived embeddings, dimensionality-agnostic representations, a meta-trained hypernetwork, multilayer perceptrons (MLPs), and retrieval within a single architecture. Pre-trained on more than 1,800 heterogeneous classification datasets, iLTM achieves consistently superior performance across tabular classification and regression tasks, from small datasets to large and high-dimensional tasks. After light fine-tuning, the meta-trained hypernetwork transfers to regression targets, matching or surpassing strong baselines. Extensive experiments show that iLTM outperforms well-tuned GBDTs and leading deep tabular models while requiring less task-specific tuning. By bridging the gap between tree-based and neural methods, iLTM offers a new framework for tabular foundation models for robust, adaptable, and scalable tabular learning.

\end{abstract}




\maketitle

\section{Introduction}
\label{introduction}

\begin{figure*}[t]
\begin{center}
\centerline{\includegraphics[width=0.95\linewidth]{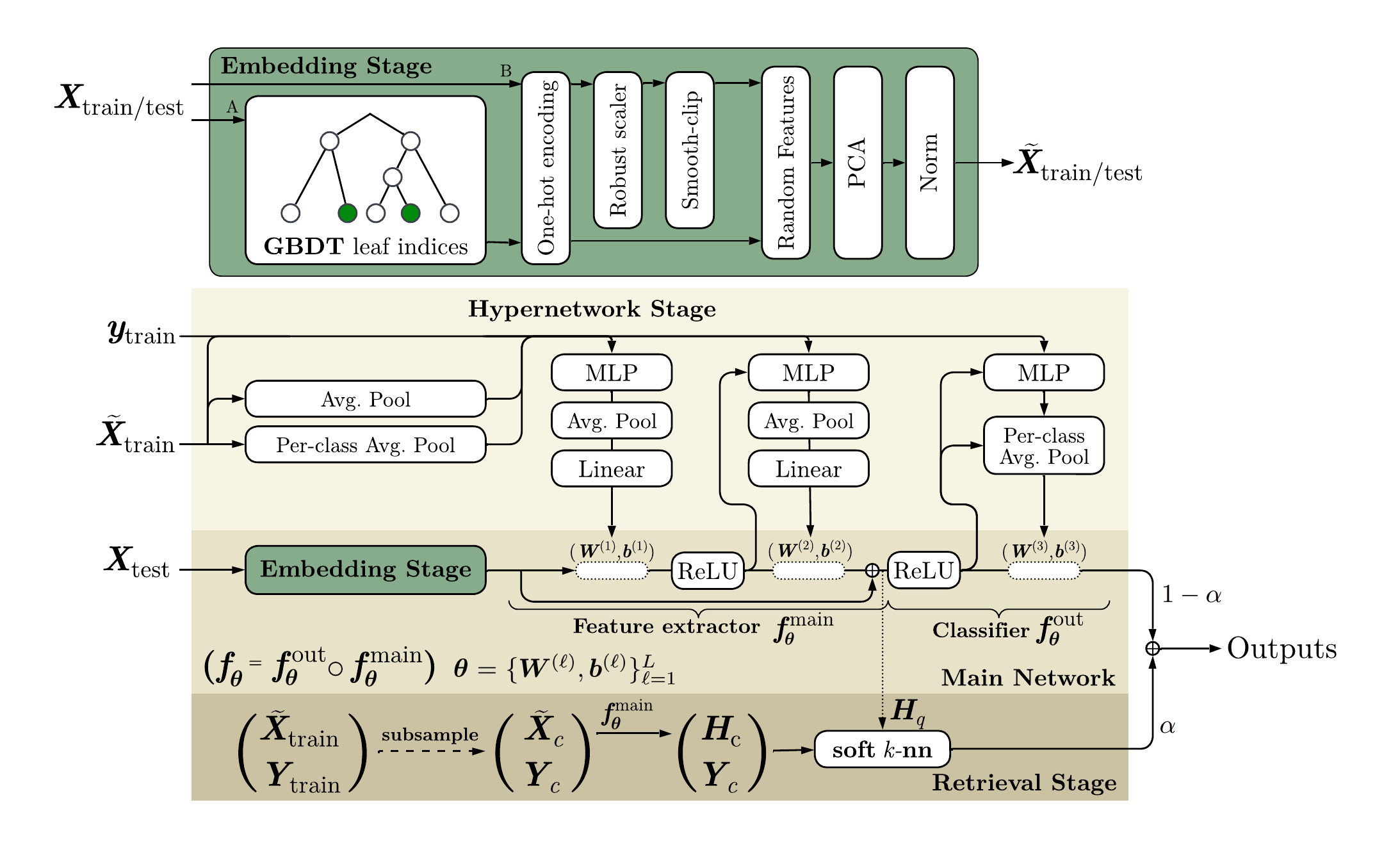}}
\Description{Diagram showing the iLTM architecture pipeline from raw tabular data through GBDT embeddings, preprocessing, hypernetwork, and main network.}
\caption{Overview of the iLTM architecture. Raw tabular datasets pass through GBDT embeddings and/or robust preprocessing, then a dimensionality-agnostic representation feeds a meta-trained hypernetwork that generates the MLP layers (3 layers, 512 hidden units) of the main network. The main network predictions are augmented with retrieval weighted by $\alpha$.}
\label{fig:iltm}
\end{center}
\end{figure*}

Tabular data is one of the most common data structures in real-world applications, including healthcare, logistics, finance, and countless administrative tasks. 
Despite its ubiquity, recent advances in large-scale foundation models that have transformed natural language processing, vision, and other domains \cite{brown2020language, rombach2022high, radford2021learning}, have not yet transferred equivalently to tabular data \cite{lu2025large}. 
Although significant progress has been made in developing neural network architectures for tabular data \cite{borisov2022deep, ye2024closer}, and in adapting existing foundation models \cite{dinh2022lift, hegselmann2023tabllm}, no single approach consistently excels across the full spectrum of tabular tasks. These tasks range from small datasets with mixed feature types to extremely large and high-dimensional tables. 
As a result, gradient-boosted decision trees (GBDTs) have remained the dominant choice for real-world tabular applications and automation tasks, largely due to their robust performance and ease of use. 
However, GBDTs and other classical models must be trained and hyperparameter-tuned independently for each new task, an approach that becomes prohibitively time-consuming in large-scale or highly diverse settings. This limitation underscores the need for a \emph{foundation model} in the tabular domain\textemdash one that is pre-trained on thousands of examples, with learned knowledge of the tabular domain, and capable of delivering strong performance across heterogeneous tasks without exhaustive re-training.

Addressing this performance gap has motivated a recent call for Large Tabular Models (LTMs) \cite{breugel2024position} to bring the success of foundation models in other domains to tabular learning.
Simple neural architectures based on multilayer perceptrons (MLPs) have been shown to be a good fit for tabular data, achieving performance comparable to GBDTs when paired with proper preprocessing \cite{holzmuller2024better} and parameter tuning \cite{kadra2021well}.
However, comprehensive evaluations like TabZilla \cite{mcelfresh2023tabzilla} reveal a critical insight: no single technique consistently excels across the diverse landscape of tabular tasks.

This observation reflects the inherent heterogeneity of tabular data, and different architectural paradigms have emerged to address specific aspects of this challenge. 
While neural networks offer flexibility and representational power \cite{borisov2022deep,ye2024closer}, they may not always capture the specific inductive biases inherent in tabular data, where GBDTs tend to excel \cite{grinsztajn2022tree, shwartz2022tabular, mcelfresh2023tabzilla}. Recent work on in-context learning for tabular data has demonstrated remarkable effectiveness in learning new tasks without re-training \cite{hollmann2023tabpfn,thomas2024localpfn,muller2025mothernet,qu2025tabicl,hollmann2025accurate,zhang2025limix,zhang2025mitra,sun2026survey}. Moreover, research on tabular meta-learning has shown how hypernetworks can be applied to pre-train on vast amounts of datasets that are not necessarily the ones used at inference \cite{bonet2024hyperfast,muller2025mothernet}, paralleling how large-scale pre-training has revolutionized deep learning models in domains such as text \cite{brown2020language}, vision \cite{radford2021learning}, and speech \cite{radford2023whisper}. Furthermore, another promising direction has been in retrieval-augmented approaches, which have shown competitive performance in many tabular datasets \cite{gorishniy2024tabr, ye2025modernnca}, though with limitations in others \cite{rubachev2025tabred}.

Recognizing that no single technique consistently outperforms others across the diverse landscape of tabular datasets, we propose an integrated approach. In this work, we introduce iLTM (\cref{fig:iltm}), an integrated Large Tabular Model that combines the strengths of gradient boosted decision trees, meta-learned hypernetworks, retrieval-augmented classification, and strong MLP architectures within a single model. 
Our experiments reveal that the same weights, meta-trained exclusively on classification, can be fine-tuned to regression targets, highlighting promising adaptation and cross-task transfer capability for tabular foundation models.
iLTM is the first integrated architecture combining GBDT embeddings, hypernetwork, retrieval and MLP with large-scale pre-training on real tabular data, showing predictive power superior to other neural methods and a stronger capacity to adapt to datasets of different sizes and configurations. 

\paragraph{Contributions.} Our main contributions are: (i) \textbf{A novel neural-tree hybrid}: We present iLTM, integrating GBDTs, hypernetworks, retrieval modules, and strong MLPs. (ii) \textbf{Large-scale meta-training}: We pre-train on thousands of real-world tabular datasets, enabling cross-task transfer to regression. (iii) \textbf{Strong performance}: iLTM consistently outperforms well-tuned GBDTs and deep tabular models across diverse benchmarks. (iv) \textbf{Robust adaptability}: Our model scales from small to high-dimensional datasets while requiring less task-specific overhead. (v) \textbf{Open-source}: We release our code and weights at \url{https://github.com/AI-sandbox/iLTM}.

\section{Related Work}
\label{related_work}

Pre-training is fundamental in machine learning \cite{brown2020language, radford2021learning}, yet its application to tabular data is recent. Early work like TabPFN \cite{hollmann2023tabpfn,hollmann2025accurate} pre-trained Transformers on synthetic data, while HyperFast \cite{bonet2024hyperfast} and XTab \cite{zhu2023xtab} scaled meta-learning to real tabular datasets. 
Other approaches include graph-based pre-training \cite{kim2024carte}, LLM fine-tuning with serialized tables \cite{hegselmann2023tabllm}, tabular diffusion models \cite{van2024latable}, and semi-supervised meta-learning \cite{nam2023stunt}. 
Parallel to pre-training, multilayer perceptrons (MLPs) with proper preprocessing and tuning remain highly competitive for tabular data \cite{kadra2021well, holzmuller2024better, ye2024closer}. While attention-based models \cite{arik2021tabnet, huang2020tabtransformer, gorishniy2021revisiting, somepalli2021saint} incorporate specialized feature interaction mechanisms, they often struggle to scale efficiently for large or high-dimensional tables compared to simpler, robust MLP baselines. 
To further enhance representations, retrieval-augmented methods incorporate relevant training examples, combining local pattern recognition \cite{bottou1992local} with learned representations \cite{gao2023retrieval, long2022retrieval}. In tabular learning specifically, methods like TabR \cite{gorishniy2024tabr} and ModernNCA \cite{ye2025modernnca} use nearest-neighbor data to refine predictions, though they can be sensitive to the quality and scalability of the retrieval index on large or diverse tables \cite{rubachev2025tabred}.  
Finally, tree ensembles like random forests \cite{breiman2001random} and GBDTs \cite{chen2016xgboost, prokhorenkova2018catboost, ke2017lightgbm} remain competitive tabular baselines due to strong inductive biases and resilience to heterogeneous data \cite{borisov2022deep, grinsztajn2022tree, shwartz2022tabular}. While several methods merge tree biases with deep networks \cite{popov20neural, ke2019deepgbm, katzir2020net, chen2024dofen, marton2024grande}, they generally lag behind optimized GBDTs on large tasks \cite{mcelfresh2023tabzilla}. 
While leveraging GBDTs as feature extractors has been shown to improve performance of probabilistic linear and neural classifiers \cite{he2014practical, ke2019deepgbm}, a unified Large Tabular Model that effectively integrates both paradigms has yet to be achieved.

\section{iLTM: Integrated Large Tabular Model}
\label{iltm}

\paragraph{Notation} We consider tabular datasets \(\cD=\{(\bx_i,y_i)\}_{i=1}^N\), where
\(\bx_i\in\R^d\) and \(y_i\in\mathcal{Y}\), partitioned into $\cD_{\text{train}}$, $\cD_{\text{test}}$, and optionally $\cD_{\text{val}}$. For classification, $\mathcal{Y} := \{1, \ldots, K\}$ where $K$ is the number of classes. In matrix form, $\bX \in \R^{N\times d}$ contains data points as rows, and the vector $\by \in \mathcal{Y}^N$ contains the corresponding labels.

\subsection{Embedding Stage}
\label{sec:embedding_stage}

In iLTM, the raw features take two possible paths that act as preprocessing and initial transformations: a) obtaining a GBDT-based embedding, described in \cref{sec:gbdt_embeddings}, and b) a preprocessing pipeline formalized in \cite{holzmuller2024better}, that one-hot encodes categorical columns, imputes missing values as zero, and robustly scales and smooth-clips all resulting columns. This combination produces well-conditioned features, which has shown to improve performance in simple MLPs without excessive sensitivity to scale or outliers.  After the initial transformations, the data representations undergo a randomized fixed-size embedding projection, described in \cref{sec:fixed_size}.

\begin{figure}[!ht]
\begin{center}
\centerline{\includegraphics[width=\columnwidth]{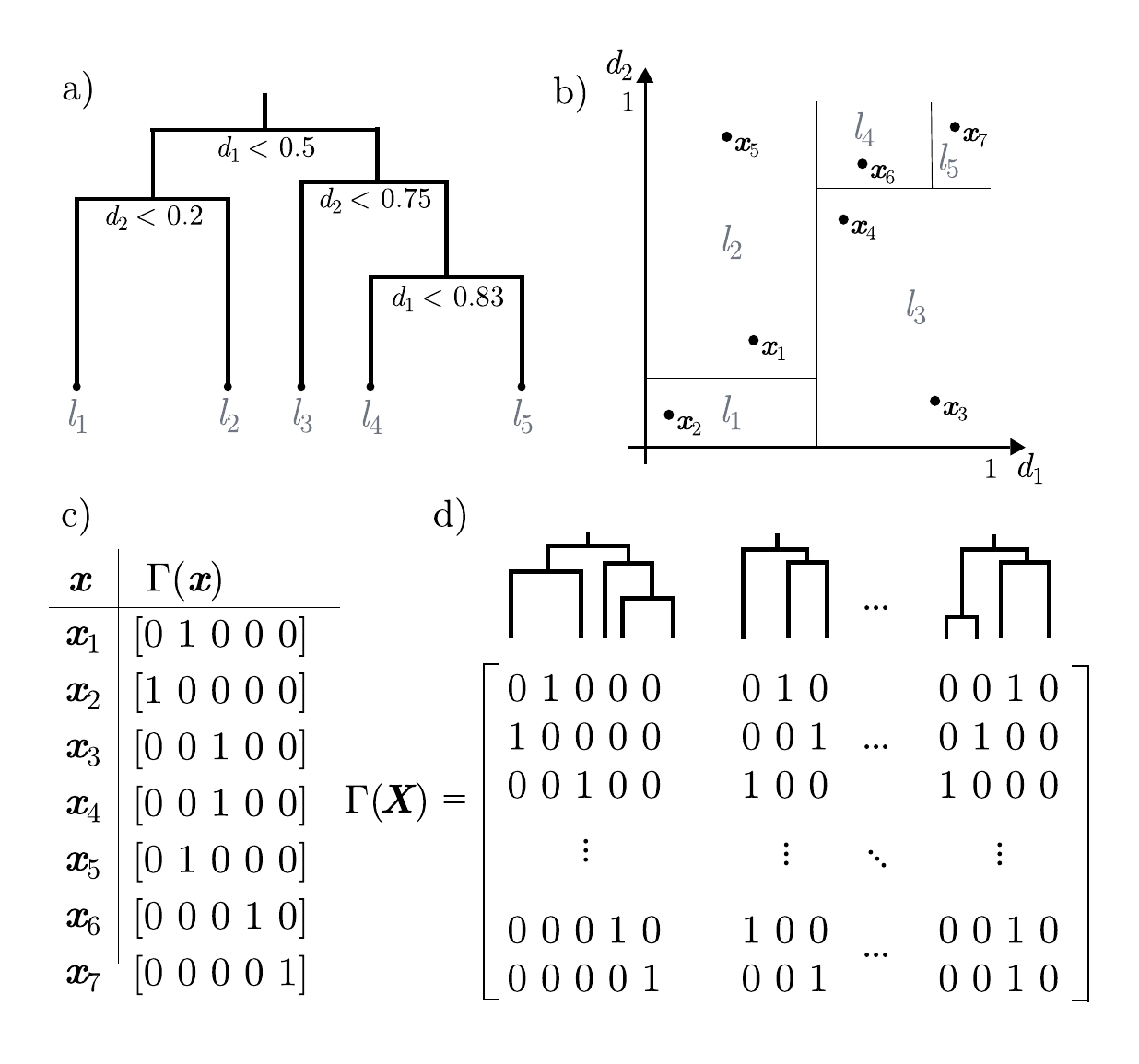}}
\caption{GBDT embedding process. (a) Decision tree splits on $d_1$ and $d_2$ leading to leaves $\{l_1, \ldots, l_5\}$. (b) 2D visualization of datapoints assigned to leaf cells. (c) Each data point is transformed into a one-hot vector. (d) Concatenated one-hot encodings across trees in the ensemble obtaining a sparse binary matrix $\Gamma(\bX)$.}
\Description{A four-part diagram showing how a decision tree splits data points into distinct leaf cells, which are then transformed into sparse, concatenated one-hot encoded vectors.}
\label{fig:tree_emb}
\end{center}
\end{figure}

\subsubsection{GBDT Embedding}
\label{sec:gbdt_embeddings}

We construct a GBDT embedding by fitting a GBDT parametrized by $\bgamma$ on a labeled set. 
Formally, let \(\{\mathrm{tree}_t\}_{t=1}^T\) be the ensemble of \(T\) decision trees, each partitioning the input space into disjoint leaf regions. However, a tree $t$ may only split along a subset of $d_t$ dimensions. 
We denote by \(L_t\) the number of leaves in \(\mathrm{tree}_t\). 
A point \(\bx_i \in \R^d\) falls into exactly one leaf for each tree, producing a leaf index \(l_t(\bx) \in \{1,\dots,L_t\}\).
We then define the GBDT embedding function \(\Gamma_{\bgamma} \colon \R^d \to \{0,1\}^M\) with \(M=\sum_{t=1}^T L_t\) by concatenating the one-hot encodings of \(l_t(\bx)\) across all \(t\in\{1,\dots,T\}\). \Cref{fig:tree_emb} illustrates the induced partitioning of the input space for a single tree into leaf regions, how data points are assigned to leaves, and how the final one-hot encoding is constructed across an ensemble, producing a high-dimensional sparse binary representation of the input data, obtaining $\Gamma_{\bgamma}(\bX) \in \{0,1\}^{N \times M}$.

This embedding incorporates the inductive biases of decision trees, and is robust to uninformative features. Additionally, some modern GBDT frameworks (such as CatBoost~\cite{prokhorenkova2018catboost} and XGBoost~\cite{chen2016xgboost}) handle categorical variables and missing values natively, improving robustness to incomplete or mixed-type datasets. 
By construction, \(\Gamma(\bx)\) is non-smooth in \(\bx\) (stepwise transitions at tree-split boundaries). 
While MLPs alone typically learn smoother functions~\cite{grinsztajn2022tree}, combining them with GBDT embeddings can capture irregular patterns more effectively. 
We empirically observed that using embeddings of non-boosted tree ensembles such as with Random Forests~\cite{breiman2001random, moosmann2006fast}, generally led to a smaller increase in performance compared to GBDT-based ones. 

Finally, we also consider concatenating \(\Gamma(\bx)\) with the features obtained after the preprocessing pipeline described in \Cref{sec:embedding_stage} to form a unified numerical-categorical representation input to the subsequent layers. 
We leave the GBDT component intentionally under-tuned, as its role is generating informative sparse embeddings rather than maximizing standalone performance, deferring predictive gains to the main network and retrieval components.

\subsubsection{Fixed-size Embedding Projection}
\label{sec:fixed_size}

We define the complete \emph{Embedding Stage} as a function \(\Psi_{\bgamma,\bomega}:\R^d\!\to \R^{d_{\text{main}}}\), that processes the raw features into a fixed-size embedding \(\tilde{\bx}\in\R^{d_{\text{main}}}\). 
Let \(\bpsi(\bx)\in\R^{m}\) denote the output of the initial transformations (either preprocessing-only, GBDT embedding, or their concatenation). 
Hence, \(\bpsi(\bx)\) may have dimension \(m\neq d\). 
We then construct \(\tilde{\bx}\) via three steps: random feature expansion, PCA-based dimensionality reduction, and feature-wise normalization. 
First, we project \(\bpsi(\bx)\) into \(\R^{r}\) via random features~\cite{rahimi2007random}:
\[
    \bpsi(\bx)\;\mapsto\;\sigma\!\bigl(\bOmega^\top\! \bpsi(\bx)\bigr), \quad \bOmega_{ij} \sim \mathcal{N}\left(0, \frac{2}{r}\right),
\]
where \(\sigma(\cdot)\) is a pointwise nonlinearity (\eg ReLU) and \(\bOmega \in \R^{m \times r}\). 
Random features can approximate certain kernels, and in this case we approximate the arc-cosine kernel, as it creates sparse, neural network-like representations~\cite{cho2009kernel}, capturing rich nonlinear behavior. We choose \(r\) large enough to preserve information but still enable subsequent efficient dimension reduction. 
Then, we reduce from \(r\) to \(d_{\text{main}}\) by applying principal component analysis (PCA) to the randomized features, 
\[
  \sigma\!\bigl(\bOmega^\top\! \bpsi(\bx)\bigr) \;\mapsto\; \bU^\top\bigl(\sigma\!\bigl(\bOmega^\top\! \bpsi(\bx)\bigr) - \bm{\mu}\bigr)\;\in\;\R^{d_{\text{main}}},
\]
where \(\bm{\mu}\) is the mean of the expanded features for the training batch and \(\bU \in \R^{r\times d_{\text{main}}}\) contains the top $d_{\text{main}}$ principal components, thus approximating Kernel PCA~\cite{lopez2014randomized}. 
We follow the procedure in~\cite{bonet2024hyperfast}, but we further normalize each \(\tilde{x}_{ij}\) by
\[
    \tilde{x}_{ij} \;\gets\; \frac{\tilde{x}_{ij} \;-\;\mu_j}{\sqrt{\sigma_j^2 + \epsilon}}.
\]
where \(\mu_j\) and \(\sigma_j\) are the mean and standard deviation of column \(j\) in a training batch \(\tilde{\bX}\). 
Thus, \(\Psi_{\bgamma,\bomega}(\bx_i)=\tilde{\bx}_i \in \R^{d_{\text{main}}}\) produces a dimensionality-agnostic embedding. Crucially, in high-dimensional settings where \(p \gg n\), neither the hypernetwork nor the main network operates directly on the raw \(p\)-dimensional input. Instead, this randomized projection compresses the feature space, decoupling the network's capacity from the raw input dimension and mitigating the curse of dimensionality prior to task-specific conditioning.

\subsection{Main Network}
\label{sec:main}
For a dataset $\cD$, a meta-model $g_{\bphi}$ generates the parameters $\btheta$ of a specialized main network $f_{\btheta}$. This main network processes inputs $\tilde{\bx} \in \R^{d_{\text{main}}}$ to obtain output logits:
\begin{equation}
    f_{\btheta}(\tilde{\bx}): \R^{d_{\text{main}}} \to \R^K
\end{equation}
The hypernetwork generates weights $\btheta$ based on the training set representations $\tilde{\bX}_{\text{train}}$ and the associated labels $\by_{\text{train}}$, allowing the main network to adapt to the specific characteristics of each task.

\subsubsection{Retrieval-Augmented Prediction}
\label{sec:retrieval}
The main network incorporates a parameter-free retrieval mechanism inspired by ModernNCA \cite{ye2025modernnca}, which operates as a soft $k$-nearest neighbors in the learned representation space \cite{cover1967nearest,goldberger2004nca}. Let $f_{\btheta} = f_{\btheta}^{\text{out}} \circ f_{\btheta}^{\text{main}}$ denote our main network, where $f_{\btheta}^{\text{main}}: \R^{d_{\text{main}}} \to \R^{d_{\text{main}}}$ outputs the penultimate layer representations and $f_{\btheta}^{\text{out}}: \R^{d_{\text{main}}} \to \R^K$ is the final classification layer. Given a query batch $\bQ \in \R^{B \times d}$ and a context set $\{\bX_c, \bY_c\}$ where $\bY_c \in \{0,1\}^{N_c \times K}$, the retrieval proceeds as follows:

\begin{enumerate}
    \item Extract penultimate layer representations:
\begin{align}
    \bH_q &= f_{\btheta}^{\text{main}}(\Psi_{\bgamma,\bomega}(\bQ)) \in \R^{B \times d_{\text{main}}}, \\
    \bH_c &= f_{\btheta}^{\text{main}}(\Psi_{\bgamma,\bomega}(\bX_c)) \in \R^{N_c \times d_{\text{main}}}.
\end{align}
\item Compute cosine similarities between query and context representations:
    \begin{equation}
        \bS = \bH_q \bH_c^\top/(\|\bH_q\|_2 \|\bH_c\|_2) \in \R^{B \times N_c}.
    \end{equation}
\item Apply temperature scaling and aggregate context labels:
    \begin{equation}
        \bA = \mathrm{softmax}(\bS/\tau) \in \R^{B\times N_c}, \qquad
\bO_{\text{retrieval}} = \bA\bY_c.
    \end{equation}
    where $\tau > 0$ controls the sharpness of the retrieval weights.
\end{enumerate}

The final class scores interpolate the main-network logits and retrieval scores:
\begin{equation}
    \bO = (1 - \alpha) f_{\btheta}^{\text{out}}(\bH_q) + \alpha \bO_{\text{retrieval}}
\end{equation}
where $\alpha \in [0,1]$ controls the retrieval contribution. When $\alpha = 0$, we have $\bO = f_{\btheta}^{\text{out}}(\bH_q)=f_{\btheta}(\bQ)$, reducing to standard prediction using only the main network. The retrieval mechanism can be activated during inference regardless of the training configuration; however, hypernetworks trained with retrieval should generate networks with embeddings better suited for similarity-based predictions.

\subsection{Hypernetwork Stage}
\label{hyper_stage}

We define the meta-model \(g_{\bphi}\) as a hypernetwork~\cite{ha2017hypernetworks} that generates the parameters \(\btheta = \{\bW^{(\ell)}, \bb^{(\ell)}\}_{\ell=1}^L\) of the main network \(f_{\btheta}\) for each dataset \(\cD\). Unlike standard approaches that train a separate model for each dataset, \(g_{\bphi}\) is trained on a collection \(\cM_{\mathrm{train}}\) of datasets and learns to produce \(\btheta\) for any new dataset at test time.

The hypernetwork $g_{\bphi}$ generates the $L$ layers sequentially, as illustrated in \cref{fig:iltm}. 
Given an embedded generation set \(\tilde{\bX}_{\text{gen}} \in\R^{N_{\text{gen}}\times d_{\text{main}}}\) and labels $\bY_{\text{gen}} \in \{0,1\}^{N_{\text{gen}}\times K}\), we obtain the global and per-class means, concatenate it with the generation set, and feed it to MLP blocks $g_{\bphi,\text{MLP}}^{(\ell)}$, consistent with the methodology proposed in \cite{bonet2024hyperfast}.
For \(\ell>1\), it additionally conditions on the previously generated layer outputs.  
We then perform on the obtained representations an average pooling over the sample dimension, producing a single dataset-level embedding \(\bz^{(\ell)} \in \R^h\). For \(\ell < L\), the hypernetwork projects \(\bz^{(\ell)}\) through a final linear layer: 
\[
  [\bW^{(\ell)}, \bb^{(\ell)}] \;=\; g_{\bphi,\text{lin}}^{(\ell)}\!\bigl(\bz^{(\ell)}\bigr).
\]
A residual connection of \(\tilde{\bX}_{\text{gen}}\) is added into the output of the penultimate layer $L-1$, preserving information from the \emph{Embedding Stage} \(\Psi\) and stabilizing hypernetwork training. 
For the last layer \(L\), let \(\bz_k^{(L)} \in \R^{d_{\text{main}}+1}\) be the average-pooled embedding for class \(k\) after $g_{\bphi}^{(L)}$. We generate \(\bW_k^{(L)}\) and \(\bb_k^{(L)}\) from \(\bz_k^{(L)}\) and form \(\bW^{(L)}\) by stacking \(\{\bW_1^{(L)}, \dots, \bW_K^{(L)}\}\) row-wise, with a similar stacking for biases \(\bb^{(L)}\). 
During meta-training, we optimize \(\bphi\) over a collection of datasets \(\cM_{\text{train}}\) to minimize:
\[
    \min_{\bphi} \E_{\cD \sim \cM_{\text{train}}}
    \Bigl[
    \cL\Bigl(
    f_{g_{\bphi}(\cD_{\text{gen}})}
    \bigl(\Psi(\cD_{\text{grad}})\bigr),
    \by_{\text{grad}}
    \Bigr)
    \Bigr],
\]
where \(\cD_{\text{gen}}\subset\cD_{\text{train}}\) and
\(\cD_{\text{grad}}\subset\cD_{\text{test}}\). 
At test time, we fix $\bphi^*$ and generate $\btheta$ for unseen tasks.

\subsection{Training iLTM}
\label{training_iltm}

Algorithm~\ref{alg:iltm_training} shows the meta-training process for iLTM. 
In practice, to avoid repeated overhead, we fit the GBDT-based embeddings and/or the robust preprocessing pipeline \emph{off-line}, storing the embedding parameters \(\bgamma\). We use both XGBoost~\cite{chen2016xgboost} and CatBoost~\cite{prokhorenkova2018catboost} libraries for GBDT implementation. 
We consider three main variants of the embedding stage: (i) GBDT-only, (ii) robust preprocessing-only, or (iii) a concatenation of both. 
These approaches are interchangeable within the same meta-training procedure and each provides different inductive biases. 
In the algorithm below, we show a simplified version where \(\bgamma\) (from GBDT or preprocessing) is already available.

For classification with \(K\) classes and labels \(\by_i\in \{0,1\}^K\), we use the standard cross-entropy loss:
\[
\cL_{\mathrm{CE}}\bigl(f_{\btheta}(\tilde{\bx}_i),\, \by_i\bigr) 
\;=\; -\sum_{k=1}^K (y_i)_k\,\log\Bigl(\mathrm{softmax}\bigl(f_{\btheta}(\tilde{\bx}_i)\bigr)_k\Bigr).
\]
The total loss is averaged over the \emph{gradient set} \(\bX_{\text{grad}}\subset\cD_{\text{test}}\), providing gradients to update \(\bphi\).

\subsection{Predicting with iLTM}
\label{predicting_iltm}

At inference time, we fix the hypernetwork \(g_{\bphi^*}\) obtained from meta-training. 
Algorithm~\ref{alg:iltm_inference} shows the deployment on a new dataset \(\cD\). 
We first fit a GBDT (or apply robust preprocessing, or both) to obtain embedding parameters \(\bgamma\). 
Then, we choose a \emph{generation subset} \((\bX_{\text{gen}}, \by_{\text{gen}})\subset\cD_{\text{new}}\) and compute \(\tilde{\bX}_{\text{gen}}=\Psi_{\bgamma}(\bX_{\text{gen}})\). 
Passing \((\tilde{\bX}_{\text{gen}}, \by_{\text{gen}})\) to the hypernetwork gives \(\btheta_{\text{new}}\), the main-network weights specialized to \(\cD_{\text{new}}\).

For query points \(\bx_*\), the model outputs \(\hat{\by}_* = f_{\btheta_{\text{new}}}\bigl(\Psi_{\bgamma,\bomega}(\bx_*)\bigr)\). 
Optionally, one may fine-tune \(\btheta_{\text{new}}\) on the training set $\cD_{\text{train}}$.
We also implement a \emph{retrieval} mechanism (cf.\ Section~\ref{sec:retrieval}), interpolating the main-network output with neighbor-based scores using a temperature \(\tau\) and weight \(\alpha\). 
Finally, an ensemble can be obtained by generating multiple \(\btheta_{\text{new}}^{(m)}\) using different generation subsets or feature bagging, and average their predictions.

\begin{algorithm}[H]
\caption{iLTM Meta-training}
\label{alg:iltm_training}
\begin{algorithmic}
\STATE {\small \textbf{Input:} Meta-collection \(\cM_{\text{train}}\), hypernetwork \(g_{\bphi}\), learning rate \(\eta\), accumulation size \(A\).}
\STATE Initialize \(\bphi\)
\FOR{iteration \(= 1\) to \text{MaxSteps}}
  \STATE \(\nabla_{\bphi} \gets 0\) \COMMENT{Reset accumulated gradient}
  \FOR{\(a = 1\) to \(A\)}
    \STATE Sample a dataset \(\cD \in \cM_{\text{train}}\)
    \STATE \textit{Load or compute} embedding parameters \(\bgamma\) (off-line)
    \STATE Choose a generation subset \((\bX_{\text{gen}}, \by_{\text{gen}})\subset\cD_{\text{train}}\)
    \STATE \(\tilde{\bX}_{\text{gen}},\, \bomega \gets \Psi_{\bgamma}(\bX_{\text{gen}})\)
    \STATE \(\btheta \gets g_{\bphi}(\tilde{\bX}_{\text{gen}},\, \bY_{\text{gen}})\) \COMMENT{Generate main weights}
    \STATE Choose a gradient subset \((\bX_{\text{grad}}, \by_{\text{grad}})\subset\cD_{\text{test}}\)
    \STATE \(\hat{\by}_{\text{grad}} \gets f_{\btheta}\bigl(\Psi_{\bgamma,\bomega}(\bX_{\text{grad}})\bigr)\)
    \STATE \(\mathcal{L} \gets \mathcal{L}_{\text{CE}}\bigl(\hat{\by}_{\text{grad}}, \by_{\text{grad}}\bigr)\) \COMMENT{Cross-entropy loss}
    \STATE \(\nabla_{\bphi} \gets \nabla_{\bphi} + \nabla_{\bphi}\,\mathcal{L}\)
  \ENDFOR
  \STATE \(\bphi \gets \bphi - \eta\,\nabla_{\bphi}\) \COMMENT{Update hypernetwork}
\ENDFOR
\STATE \textbf{return} \(\bphi\)
\end{algorithmic}
\end{algorithm}

\section{Implementation}

\subsection{Meta-Training Data Collection}

We construct a meta-training collection from approximately 5{,}000 publicly available classification datasets on OpenML~\cite{feurer2021openml, bischl2025openml}. To prevent data leakage, we apply strict discarding criteria (fully detailed in \cref{sec:discarding-criteria}) to remove any dataset overlapping with or highly similar to our evaluation benchmarks, retaining 1{,}806 valid datasets.

\subsection{Model}

For the embedding stage of the iLTM model (described in detail in \cref{sec:embedding_stage}), we use random features of dimension \(r = 2^{15}\) followed by a principal component reduction to dimension \(d_{\mathrm{main}}=512\) ~(\cref{sec:fixed_size}). 
Dimensions of 1024 were computationally prohibitive and required $\gg$24 GB VRAM, while 256 significantly degraded predictive performance. During pre-training we fix the maximum number of boosting rounds in the GBDT embedding to 100, and apply early stopping, which balances expressivity, run-time, and embedding dimension, preventing feature explosion. 
The hypernetwork of iLTM is built from MLP blocks with 1024 hidden units, followed by average pooling over the sample dimension to generate each layer’s parameters in the main network. The main network \(f_{\boldsymbol{\theta}}(\tilde{\boldsymbol{x}})\) is an MLP with 512 hidden units in each layer. We fix a batch size of 2048 for meta-training. This configuration yields strong performance while running in a single A5500 GPU with 24 GB.

\subsection{Pre-Training}
\label{appendix_pre-training}

We follow the meta-training procedure in Algorithm~\ref{alg:iltm_training} (Section~\ref{training_iltm}) on a collection of 1806 tabular datasets. To avoid repeatedly fitting GBDTs online, we train each GBDT (or robust preprocessing) offline for each dataset and cache the resulting transformations. This procedure yields \(\Gamma_{\boldsymbol{\gamma}}(\boldsymbol{X}) \in \{0,1\}^{N \times M}\) (if using GBDT embeddings) or the appropriately scaled and encoded features (if using robust preprocessing). We then apply the random features and PCA to obtain \(\tilde{\boldsymbol{X}}_{\text{gen}}\). 

\begin{algorithm}[H]
\caption{Inference with iLTM}
\label{alg:iltm_inference}
\begin{algorithmic}[1]
\STATE \textbf{Input:} Fixed hypernetwork \(g_{\bphi^*}\), new dataset \(\cD\).
\STATE Fit GBDT and/or preprocessing on \(\cD_{\text{train}}\) to obtain \(\bgamma\).
\STATE Select a generation subset \((\bX_{\text{gen}}, \by_{\text{gen}})\subset \cD_{\text{train}}\).
\STATE \(\tilde{\bX}_{\text{gen}},\, \bomega \gets \Psi_{\bgamma}(\bX_{\text{gen}})\).
\STATE \(\btheta_{\text{new}} \gets g_{\bphi^*}\bigl(\tilde{\bX}_{\text{gen}},\, \bY_{\text{gen}}\bigr)\).
\STATE \textbf{For} any sample \(\bx_*\in\cD_{\text{test}}\):
\[
   \hat{\by}_* \gets f_{\btheta_{\text{new}}}\Bigl(\Psi_{\bgamma,\bomega}(\bx_*)\Bigr).
\]
\STATE {\small \textbf{Optionally} incorporate retrieval-based predictions with weight \(\alpha\) and temperature \(\tau\).
\STATE \textbf{Optionally} fine-tune \(\btheta_{\text{new}}\) with \(\cD_{\text{train}}\).
\STATE \textbf{Optionally} repeat Steps 3--6 with different \((\bX_{\text{gen}}, \bY_{\text{gen}})\) sample or feature subsets to form an ensemble; average predictions.
\STATE \textbf{return} \(\hat{\by}_{\text{new}}\).}
\end{algorithmic}
\end{algorithm}

Although one could refit multiple GBDTs per dataset with different seeds to increase data diversity, we found such augmentation computationally prohibitive. Instead, each dataset is loaded with its fixed embedding during the meta-training loop. 
Our meta-training runs typically converge within $\sim$400\,000 steps with a gradient accumulation size $A=40$, after which we select the checkpoint that yields the best meta-validation performance (e.g.\ few-shot accuracy on held-out datasets).

\subsection{Inference}
\label{appendix_inference}

At inference on a new dataset \(\mathcal{D}\), we first fit the chosen embedding \(\Gamma_{\boldsymbol{\gamma}}\) or robust scaling on \(\mathcal{D}_{\mathrm{train}}\). We then choose a generation subset \(\{\boldsymbol{X}_{\mathrm{gen}}, \boldsymbol{y}_{\mathrm{gen}}\} \subset \mathcal{D}_{\mathrm{train}}\). For small \(\mathcal{D}_{\mathrm{train}}\), we often use all samples, whereas for large \(\mathcal{D}_{\mathrm{train}}\), we sample a subset. After mapping \(\boldsymbol{X}_{\mathrm{gen}}\) to \(\tilde{\boldsymbol{X}}_{\mathrm{gen}}\), we feed \(\tilde{\boldsymbol{X}}_{\mathrm{gen}}\) and \(\boldsymbol{y}_{\mathrm{gen}}\) into \(g_{\boldsymbol{\phi}}^*\) to generate \(\boldsymbol{\theta}_{\mathrm{new}}\). Any test point \(\boldsymbol{x}_*\in\mathcal{D}_{\mathrm{test}}\) is mapped to \(\tilde{\boldsymbol{x}}_*\) and passed through the main network to predict logits \(f_{\boldsymbol{\theta}_{\mathrm{new}}}(\tilde{\boldsymbol{x}}_*)\). 

We allow a \emph{dynamic} strategy for fitting the GBDT embedding: if the dataset size is under 2\,000 samples, we use all training points to fit the GBDT. Otherwise, we use 50\% of the training data, up to a maximum size of 100\,000 samples, and use the rest to sample the generation sets for the hypernetwork. All GBDTs are trained with early stopping (with 50 rounds without improvement) to mitigate overfitting and reduce run-time. This integrated inference procedure allows iLTM to adapt flexibly to various data scales while preserving strong predictive performance.

Optionally, one may fine-tune \(\boldsymbol{\theta}_{\mathrm{new}}\) on \(\mathcal{D}_{\mathrm{train}}\). We also implement feature bagging and multi-subset ensembling by generating multiple \(\boldsymbol{\theta}_{\mathrm{new}}^{(m)}\) from different subsets \(\{\boldsymbol{X}_{\mathrm{gen}}^{(m)}, \boldsymbol{y}_{\mathrm{gen}}^{(m)}\}\) and averaging their predictions. 

\section{Benchmarks}

\subsection{TabZilla Hard Benchmark}
\label{sec:benchmark}

To compare iLTM to existing machine learning models, we evaluate it on the TabZilla Hard Benchmark Suite \cite{mcelfresh2023tabzilla}. This comprises 36 datasets specifically selected for baseline resistance, algorithmic sparsity (few models achieve top performance), and GBDT underperformance, ensuring highly diverse algorithmic challenges.

\begin{figure*}[!ht]
    \centering
    \includegraphics[width=1\linewidth]{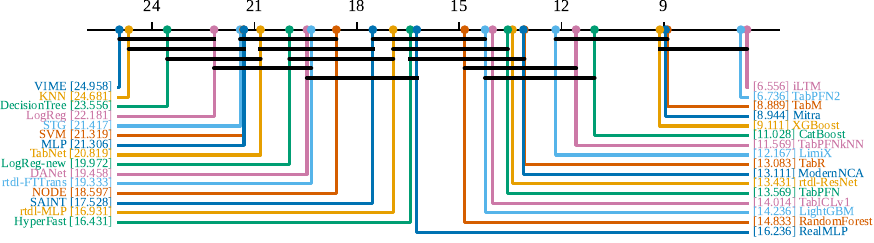}
    \caption{Critical difference diagram \cite{demvsar2006statistical} of average AUC rankings, showing algorithm groups determined by the Conover post-hoc test \cite{conover1979multiple} following a Friedman test \cite{friedman1937use} at significance level 0.05. For algorithms with missing runs, their average rank was imputed using their performance on completed datasets, while those with more than ten missing datasets were removed. Note that iLTM and TabPFNv2 (among others) ran on all datasets.}
    \Description{A Critical Difference diagram ranking various machine learning algorithms by average AUC. iLTM is clustered in the top-performing group.}
    \label{fig:conover_cd_mean_rank}
\end{figure*}

We follow the TabZilla benchmarking pipeline for both benchmark sets \cite{mcelfresh2023tabzilla}, adhering to their test folds, preprocessing steps, and evaluation metrics and constraints. In particular, each model is run once with default hyperparameters, and 29 additional ones with random hyperparameters. Each run can take up to 2 hours, and the total runtime cannot exceed 10 hours. In contrast to our lightweight GBDT embedding stage, GBDT baselines like XGBoost and CatBoost on tabular benchmarks are typically run with extensive hyperparameter sweeps and, on larger datasets, often fit an order of magnitude more trees than we do.

\begin{figure*}[!t]
    \centering
    \includegraphics[width=\textwidth]{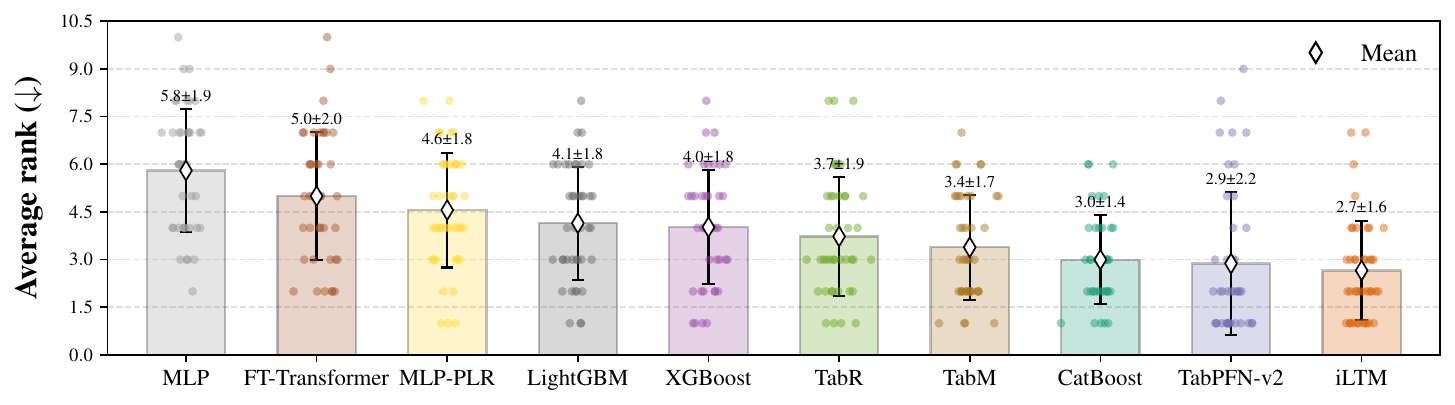}
    \caption{Average rank ($\downarrow$) on the 18 public regression datasets from \cite{grinsztajn2022tree} also used in \cite{gorishniy2024tabr, gorishniy2025tabm}. The meta-trained iLTM, fine-tuned on each task, achieves the top overall rank, outperforming gradient-boosted trees and performing on par with recent deep tabular models, illustrating that features learned during classification pre-training transfer effectively to regression problems.}
    \Description{A bar chart showing the average rank of various models on 18 regression datasets. iLTM has the lowest and best average rank, outperforming baselines like TabM, TabPFNv2, CatBoost, and XGBoost.}
    \label{fig:regression_rank}
\end{figure*}

Using the public results from \cite{mcelfresh2023tabzilla}, we compare iLTM with all TabZilla methods that successfully ran on the vast majority of datasets by excluding only methods that failed on 10 or more datasets. The resulting pool of included original baselines is: $k$-nearest neighbors (KNN) \cite{cover1967nearest}, logistic regression \cite{cox1958regression}, decision trees \cite{quinlan1986induction}, Random Forests \cite{breiman2001random}, SVMs \cite{cortes1995support}, VIME \cite{yoon2020vime}, STG \cite{yamada2020feature}, TabNet \cite{arik2021tabnet}, DANets \cite{chen2022danets}, NODE \cite{popov20neural}, MLP (rtdl-MLP), ResNet (rtdl-ResNet), FT-Transformer (rtdl-FTTrans) \cite{gorishniy2021revisiting}, SAINT \cite{somepalli2021saint}, LightGBM \cite{ke2017lightgbm}, TabPFN \cite{hollmann2023tabpfn}, CatBoost \cite{prokhorenkova2018catboost}, and XGBoost \cite{chen2016xgboost}. In addition, we extend the benchmark by introducing several additional baselines: LogReg-new, which extends the original logistic regression setup \cite{cox1958regression} with a larger hyperparameter search, and several newer methods, namely, HyperFast \cite{bonet2024hyperfast}, TabR \cite{gorishniy2024tabr}, RealMLP \cite{holzmuller2024better}, ModernNCA \cite{ye2025modernnca}, TabM \cite{gorishniy2025tabm}, TabPFN-kNN \cite{thomas2024localpfn}, TabICL \cite{qu2025tabicl}, TabPFNv2 \cite{hollmann2025accurate}, LimiX \cite{zhang2025limix}, and Mitra \cite{zhang2025mitra}. We note that HyperFast shares several architectural similarities with iLTM and, unlike iLTM, may have been pre-trained on datasets that overlap with the benchmark suite. For TabPFN \cite{hollmann2023tabpfn}, we subset to 30 features that maximize mutual information, select 3000 random samples, and use one-vs-rest classification, as the model does not support larger datasets or more than 10 classes. Similarly, for TabPFNv2 \cite{hollmann2025accurate}, we subset to the maximum supported limit of 500 features and 10{,}000 samples, and use one-vs-rest for more than 10 classes.

Our results demonstrate that iLTM obtains the best mean AUC ranking among all compared methods, as shown in the critical difference diagram (\cref{fig:conover_cd_mean_rank}). The diagram reveals several statistically indistinguishable groups of algorithms according to the Conover post-hoc test, with iLTM, TabPFNv2, TabM, Mitra, and XGBoost forming the top-performing group. The search space of hyperparameters used on TabZilla for iLTM and all added baselines, additional metrics, and complete results for the top methods on TabZilla Hard can be found in \cref{appendix_benchmarks}.

\subsection{High-Dimensional Collection Benchmark}

For our curated high-dimensional dataset collection, mostly composed of biomedical datasets, we focus our comparison on XGBoost \cite{chen2016xgboost} as a close competitor that ran on all the datasets for TabZilla Hard \cite{mcelfresh2023tabzilla}. This collection spans tasks with up to 7{,}000 samples and 19{,}993 features, allowing us to specifically test performance in settings where the number of features can be extremely large relative to the number of observations. Note that TabPFNv2 \cite{hollmann2025accurate} is not designed for large tables, and TabR \cite{gorishniy2024tabr} suffered out-of-memory or time-limit errors on several of the TabZilla Hard datasets that are large. For these experiments, we follow the TabZilla setup with the same compute and time constraints (\cref{sec:benchmark}). \cref{tab:auc_highdim} shows how iLTM maintains its performance advantage on high-dimensional tasks, achieving a higher average AUC compared to XGBoost. The results indicate that iLTM outperforms XGBoost on the majority of the evaluated datasets, with notable improvements on biomedical applications such as SMK-CAN-187 and TOX-171. More details about these benchmarks are available in \cref{appendix_highdim}.

\begin{table}[ht]
\centering
\caption{AUC scores for high-dimensional datasets.}
\begin{tabular}{lcc}
\toprule
Dataset & XGBoost & iLTM \\
\midrule
CLL-SUB-111 & $0.9317_{\pm 0.0516}$ & $0.9577_{\pm 0.0405}$ \\
lung & $0.9633_{\pm 0.0644}$ & $0.9897_{\pm 0.0182}$ \\
Prostate-GE & $0.9360_{\pm 0.1295}$ & $0.8967_{\pm 0.1295}$ \\
SMK-CAN-187 & $0.6890_{\pm 0.1779}$ & $0.7336_{\pm 0.0943}$ \\
TOX-171 & $0.9377_{\pm 0.0277}$ & $0.9750_{\pm 0.0269}$ \\
arcene & $0.8921_{\pm 0.0812}$ & $0.8831_{\pm 0.0812}$ \\
gisette & $0.9970_{\pm 0.0013}$ & $0.9976_{\pm 0.0015}$ \\
\midrule
Average & $0.9067_{\pm 0.0762}$ & $\mathbf{0.9191_{\pm 0.0567}}$ \\
\bottomrule
\end{tabular}
\label{tab:auc_highdim}
\end{table}

\subsection{Transfer to Regression Tasks}

As shown in Fig.~\ref{fig:regression_rank}, despite being pre-trained solely on classification corpora, iLTM transfers smoothly to regression problems after short fine-tuning. 
It obtains the best average rank on the 18 public regression datasets from \cite{grinsztajn2022tree} also used in recent works \cite{gorishniy2024tabr, gorishniy2025tabm}, outperforming GBDT models and recent deep tabular methods like TabM and TabPFNv2.

To apply iLTM to regression tasks, the same architecture is used and per-class operations become dataset-level where needed. 
After the initial transformation, we compute the global mean of the transformed features over the batch, replacing the per-class means used in classification. 
The standardized target $\by$ of the batch is passed to the hypernetwork, instead of a one-hot label tensor.   
The hypernetwork block processes the embeddings exactly as in classification, generating a set of layer weights. 
For the final prediction layer, the per-sample weight predictions are averaged to obtain a single weight vector, which is reshaped into a linear mapping that produces the regression output. 

\begin{table}[h]
    \caption{Performance (RMSE) improvement and fine-tuning time of iLTM on regression tasks compared to starting from random initialization of main network.}
    \label{reg_ablation}
    \centering
    \begin{tabular}{lcc}
    \toprule
    \textbf{\small Configuration} & \multicolumn{1}{c}{\textbf{\small Improvement}} & \multicolumn{1}{c}{\textbf{\small Time (s)}} \\
    \midrule
    Random Init. (Baseline) & -- & $47.151\ \pm\ 9.502$ \\
    Hypernetwork (Transfer) & $46.2\%\ \pm\ 8.4\%$ & $24.348\ \pm\ 5.961$ \\
    Hypernetwork + Ens. ($\times5$) & $51.0\%\ \pm\ 7.4\%$ & \;$\,97.271\ \pm\ 25.467$ \\
    \bottomrule
    \end{tabular}
\end{table}

To obtain accurate predictions for regression, some fine-tuning steps are required because the hypernetwork was pre-trained on classification tasks. 
\cref{reg_ablation} shows the average performance improvement when using the weights generated by iLTM and fine-tuning for regression tasks, compared to randomly initializing the linear layers (no use of the pre-trained hypernetwork) on the validation set of 5 datasets from \cite{gorishniy2024tabr} (\emph{california}, \emph{diamond}, \emph{isolet}, \emph{analcatdata supreme}, \emph{fifa}). 
Not only is improvement significant, but fine-tuning time is halved compared to training from scratch. 
Ensembling further boosts performance, and training five predictors only doubles the time relative to a single randomly initialized network. 
This indicates that the representations learned during classification pre-training can generalize well across task types, enabling iLTM to deliver state-of-the-art performance with minimal adaptation.

\subsection{TabReD Benchmark}

Finally, we evaluate iLTM on TabReD \cite{rubachev2025tabred}, a benchmark with industry-grade tabular datasets and time-based splits that includes both classification and regression tasks. The benchmark contains large-scale real-world tables, with the evaluated subsets reaching up to 423{,}000 samples and 986 features, and several tasks originating from full datasets with tens of millions of observations. We include models used in real-world production settings that can scale to the size of the TabReD datasets \cite{wang2021dcn, anil2022factory}, classical GBDT models \cite{chen2016xgboost, prokhorenkova2018catboost, ke2017lightgbm}, MLP-based models \cite{gorishniy2021revisiting, klambauer2017self, kirichenko2023last, gorishniy2022embeddings, gorishniy2025tabm}, a Transformer-based method \cite{gorishniy2021revisiting}, and a retrieval-based method \cite{gorishniy2024tabr}. iLTM obtains very competitive results and surpasses GBDT-only methods and recent deep tabular baselines, on both regression and classification tasks, obtaining the best average rank across the 8 tasks. The complete results on the TabReD benchmark are shown in \cref{appendix:tabred_results}.

\subsection{Weight Space Analysis}

Beyond the main benchmark evaluations, \cref{sec:wsl} explores the weight space structure induced by the meta-trained hypernetwork, revealing domain-based task clustering and increased ensemble diversity during fine-tuning. 

\subsection{Few-Shot Meta-Validation}
\label{sec:fewshot}

\begin{figure}[h]
    \centering
    \includegraphics[width=\linewidth]{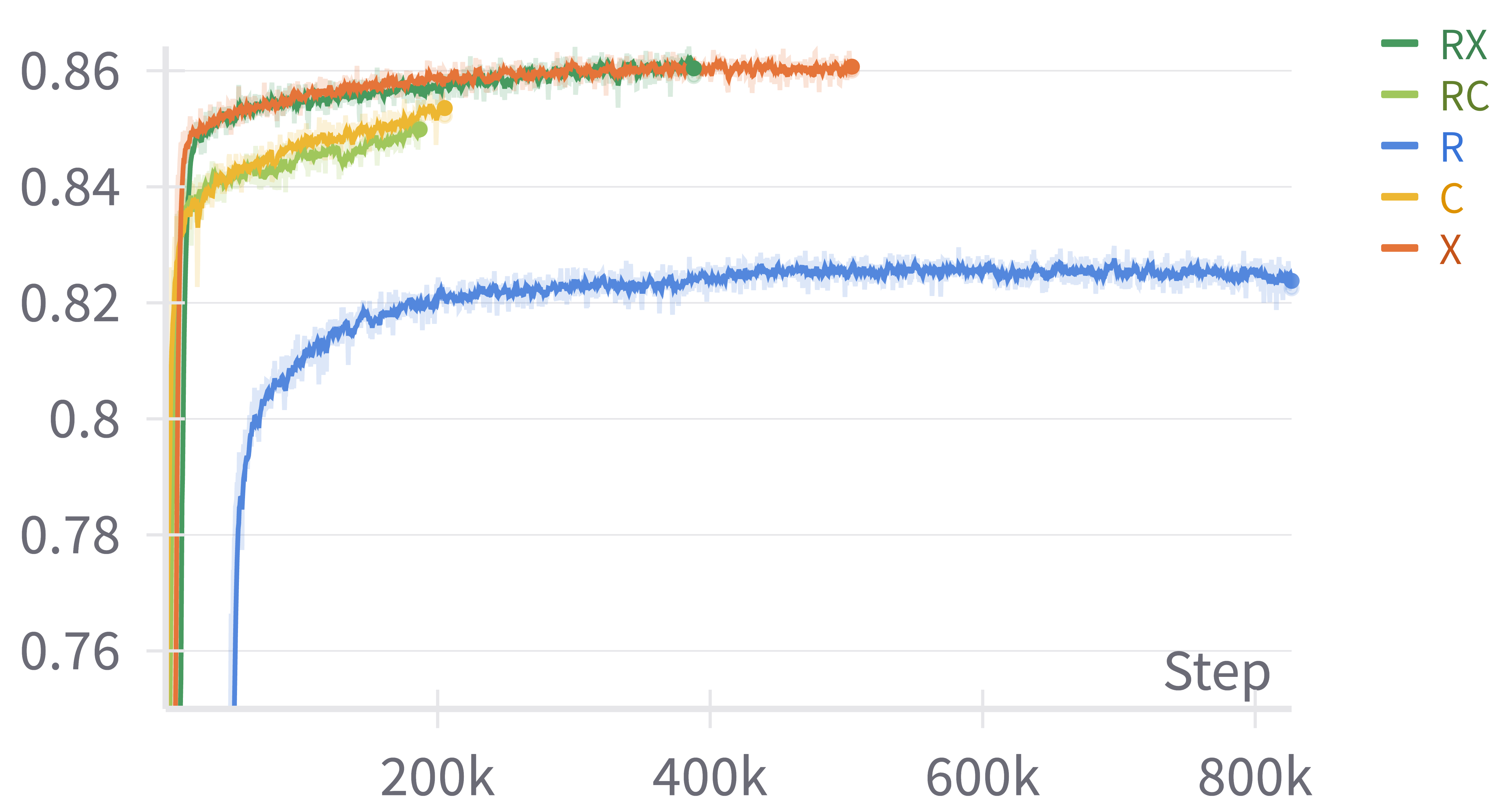}
    \caption{Average few-shot meta-validation AUC for each preprocessing: R (robust without GBDT), X (XGBoost-based), C (CatBoost-based), and the concatenations RX and RC.}
    \Description{A line graph tracking average few-shot meta-validation AUC over training steps for five different preprocessing configurations.}
    \label{fig:wandb}
\end{figure}

\cref{fig:wandb} compares performance across five preprocessing configurations during meta-validation in pre-training. 
Because R-only runs complete each training step more quickly, they can accumulate the highest number of updates in the seven-day limit, with X next in throughput and C the slowest. 
Including GBDT embeddings (X or C) yields higher few-shot accuracy early on, an effect clearly visible when validating on a single batch without fine-tuning, yet this comes at the cost of slower iteration. 
Concatenations RX and RC also show similarly high few-shot gains, but run marginally slower than single-embedding configurations, as they process both sets of features. 
Notably, GBDT-based embeddings provide a strong initialization for new tasks, which helps explain their advantage in a few-shot setting. 
Nonetheless, once full fine-tuning is allowed over one whole dataset during inference, R-only can match or even surpass the X- and C-based variants despite its initially lower few-shot performance.

\subsection{Core Ablation Study}
\label{sec:core_ablation}

This section presents an ablation study to demonstrate the effectiveness of the pre-trained hypernetwork within the iLTM framework.
The experiments are conducted on the \emph{cylinder-bands}, \emph{ecoli}, \emph{connect-4}, \emph{christine}, and \emph{volkert} datasets from OpenML \cite{OpenML2013, bischl2025openml}, selected for their varied dimensions.
The pre-specified data folds are used for evaluation, and all executions are performed on a single NVIDIA A5500 GPU. 

Table \ref{ablation_hypernetwork} presents a comparative performance analysis across three configurations: the main network whose weights are generated by iLTM's hypernetwork in a single forward pass from a batch of data, the iLTM-generated main network with subsequent fine-tuning with early stopping on the complete training set, and a main network with randomly initialized weights trained from scratch with early stopping on the complete training set.
The performance is assessed based on AUC, fit time (in seconds), and prediction time (in seconds).

\begin{table}[!ht]
    \caption{Performance comparison demonstrating the effectiveness of the iLTM hypernetwork.}
    \label{ablation_hypernetwork}
    \centering
    \setlength{\tabcolsep}{4.5pt}
    \begin{tabular}{lccc}
    \toprule
    \textbf{\small Config} & \textbf{\small AUC} & \textbf{\small Fit time (s)} & \textbf{\small Pred. time (s)} \\
    \midrule
    Random Init. & $0.826_{\pm0.167}$ & $94.013_{\pm247.115}$ & $0.969_{\pm0.810}$ \\\midrule
    iLTM (1 Batch) & $0.803_{\pm0.129}$ & $3.292_{\pm3.588}$ & $0.955_{\pm0.802}$ \\
    iLTM + Fine-tune & $0.856_{\pm0.134}$ & $27.229_{\pm32.830}$ & $0.949_{\pm0.791}$ \\
    \bottomrule
    \end{tabular}
\end{table}

The results presented in Table \ref{ablation_hypernetwork} clearly demonstrate the significant advantages of employing the pre-trained hypernetwork in iLTM. 
Regarding AUC performance, the iLTM with fine-tuning configuration achieves the highest AUC, outperforming the standard approach of training a neural network from scratch. 
This indicates that the hypernetwork provides a superior weight initialization, leading to a better final model after fine-tuning. 
Even the configuration using iLTM without fine-tuning yields a respectable AUC. 
While lower than the fine-tuned models, this result highlights that the hypernetwork can generate reasonably effective main network weights directly.

In terms of fit time, the most striking benefit is observed.
iLTM base is exceptionally fast, as it only requires a single forward pass on one batch of data through the hypernetwork.
Crucially, when adding fine-tuning, it also requires significantly less time to converge compared to training from random weights.
The large standard deviation in fit time for training from random initialization also suggests that training from scratch can be more variable and prone to longer convergence times.
Prediction times are comparable across all three configurations, as the underlying architecture is the same.

\begin{figure}[ht]
    \centering
    \includegraphics[width=0.9\linewidth]{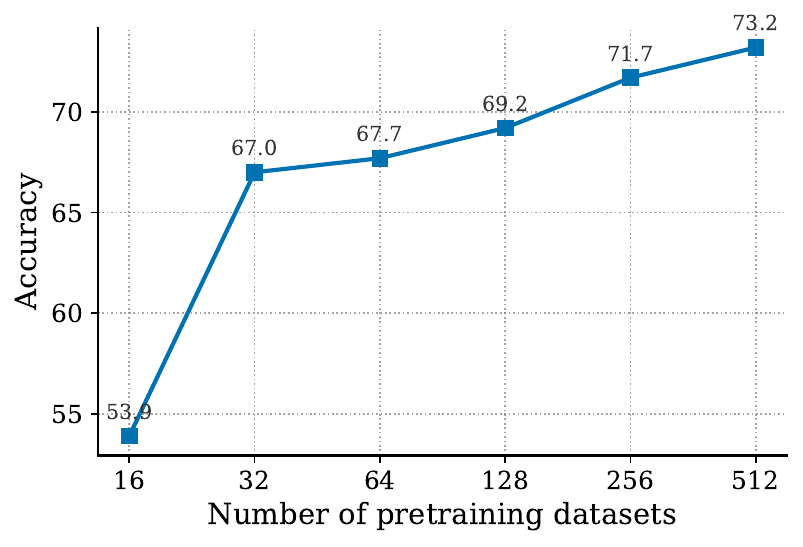}
    \caption{iLTM average meta-validation accuracy scaling as a function of pre-training dataset count.}
    \Description{A line graph plotting iLTM average meta-validation accuracy on the vertical axis against the number of pre-training datasets on the horizontal axis. The line shows a clear and consistent upward trend as the dataset count increases from 16 up to 512.}
    \label{fig:scaling}
\end{figure}

As shown in Figure~\ref{fig:scaling}, predictive performance scales consistently with the number of pre-training datasets when evaluated under a fixed budget.

In conclusion, these findings underscore that the hypernetwork in iLTM provides a dual advantage: it initializes the main network with weights that lead to higher AUC after fine-tuning compared to random initialization, and it dramatically reduces the fit time needed to achieve optimal or near-optimal performance.  

We also include ablation studies isolating the contributions of ensembling, retrieval, and fine-tuning hyperparameters in \cref{appendix_ablation_hyperparameters}.

\section{Discussion}
\label{discussion}

We introduce iLTM, a large tabular model that integrates tree-based representations, meta-learned hypernetworks, and deep neural architectures with retrieval-augmented predictions into a single model that yields strong and reliable performance across a diverse range of tabular tasks. 
By incorporating gradient-boosted decision-tree embeddings, our method inherits the robust inductive biases that have long made tree ensembles the preferred choice for many real-world tabular problems \cite{breiman2001random, chen2016xgboost, prokhorenkova2018catboost, ke2017lightgbm, borisov2022deep, grinsztajn2022tree}.
At the same time, the hypernetwork-based MLP backbone and retrieval module provide a scalable and expressive framework that benefits from large-scale meta-training while remaining flexible enough to adapt quickly to new tasks, and they connect to retrieval-augmented modeling that has proven effective in other domains and in tabular settings \cite{gorishniy2024tabr, ye2025modernnca}. 
Crucially, our experiments demonstrate that this neural-tree hybrid approach not only matches but surpasses GBDTs and well-tuned neural baselines, minimally tuning the GBDT embedding, suggesting that both the tree-based and neural paradigms contribute complementary strengths within the iLTM architecture.

Strong performance of iLTM spans very different tabular distributions and scales. 
On relatively small datasets with mixed feature types, GBDT embeddings effectively capture discrete feature interactions, which are then refined by the MLP. 
Meanwhile, in large and high-dimensional settings, iLTM’s neural capacity and retrieval mechanism allow it to scale more gracefully than methods that rely on massive ensembles or carefully tuned attention blocks \cite{gorishniy2021revisiting, huang2020tabtransformer}. 

This adaptability partly arises from the meta-trained hypernetwork, which internalizes knowledge from a large and varied corpus of tabular datasets. 
By distilling patterns that frequently arise across heterogeneous tasks, the hypernetwork can generate MLP weights that give a favorable initialization, requiring minimal fine-tuning at test time. 
While standard GBDTs often demand fresh hyperparameter searches for each new dataset, iLTM amortizes the effort of meta-training, cutting down the hyperparameter optimization for the GBDT part, and deployment time without sacrificing accuracy. 

From a practical standpoint, the inclusion of GBDT embeddings introduces an additional step that adds overhead relative to a purely neural pipeline. However, our analysis reveals a regime separation whose optimal configuration is dataset- and budget-dependent. As shown during meta-validation (\cref{fig:wandb}), GBDT embeddings generally provide a stronger initialization for rapid, single-forward-pass inference across the broader task distribution by mapping features into a structured sparse binary format. Conversely, the purely neural robust-preprocessing variant avoids tree-fitting overhead entirely. While its single-pass initialization varies by dataset, it consistently matches or surpasses the GBDT embeddings once full dataset fine-tuning is computationally permissible. Thus, iLTM offers a flexible deployment spectrum rather than a rigid pipeline.

Being pre-trained exclusively on classification tasks, iLTM shows strong transfer learning capabilities on regression tasks, but performance on other out-of-distribution tasks could be limited.
Second, fitting a GBDT for every new dataset introduces an extra processing stage that can be prohibitive in latency-critical scenarios. However, this step is optional, and the robust preprocessing can be used without fitting any GBDT. Finally, the retrieval module employs a fixed similarity metric; when the feature distribution drifts, this static metric can select sub-optimal neighbors, limiting the benefits of retrieval-augmented inference \cite{rubachev2025tabred}. 

Future directions could explore extending the meta-training corpus to include regression and other tasks to broaden the model's inductive biases. Making the retrieval component more dynamic by learning an adaptive similarity metric might further improve accuracy on niche domains where local neighborhoods offer important information, \eg precision health or finance. Incorporating lightweight attention blocks directly into the main network could enhance the capture of higher-order interactions not explicitly represented in tree splits. Finally, further probing the internal structure of the hypernetwork-generated weight space $\theta\in\Theta$ \cite{schurholt2024towards} may provide deeper insights into how iLTM generalizes across tasks and inform both compression and interpretability efforts.

Overall, the results advocate for a paradigm shift that unifies traditionally separate lines of research in tabular learning. By demonstrating that tree embeddings, meta-learning, and deep neural methods can be fused into a cohesive \emph{foundation model} approach, iLTM paves the way for future tabular models that seamlessly adapt to a wide array of settings with limited tuning, bridging the gap between specialized methods for small, structured datasets and the large-scale neural approaches that dominate other domains.

\bibliographystyle{ACM-Reference-Format}
\bibliography{refs}

\appendix

\section{Extended Notation}
\label{appendix_iltm_notation}

To ease navigation, \cref{tab:notation} compiles the notation used in the paper. 
Lower-case bold letters (e.g.\ $\bx$) denote vectors, upper-case bold
(e.g.\ $\bX$) denote matrices, and calligraphic (e.g.\ $\cD$) denote
sets unless noted otherwise (e.g., $\cL$ denotes a loss function). Readers may find it convenient to keep this table at hand while reading \cref{iltm}.

\begin{table}[!ht]
\centering
\small
\caption{Notation table.}
\begin{tabular}{p{0.17\columnwidth} p{0.73\columnwidth}}
\toprule
\textbf{Notation} & \textbf{Description} \\
\midrule
$\cD$ & A dataset. \\
$\cD_{\text{train}}$, $\cD_{\text{test}}$, $\cD_{\text{val}}$ & Training, test, and validation splits of $\cD$. \\
$\cM_{\text{train}}$, $\cM_{\text{test}}$, $\cM_{\text{val}}$ & Meta-training, meta-test, and meta-validation collections of datasets. \\
$N$ & Number of samples in a dataset. \\
$d$ & Number of raw features in the data. \\
$\bx$ & A single data point. \\
$\bX $ & Data matrix where each row is a data point. \\
$K$ & Number of classes in the classification task. \\
$\mathcal{Y}$ & $\{1, \ldots, K\}$ \\
$\by$ & Label vector for the data. \\
$\bY$ & Label matrix (one-hot encoded) for the data. \\
$\bgamma$ & Parameters of the GBDT and preprocessing pipeline. \\
$\Gamma_{\bgamma}(\cdot)$ & GBDT embedding function parameterized by $\bgamma$. \\
$T$ & Number of trees in the GBDT ensemble. \\
$L_t$ & Number of leaves in tree $t$. \\
$l_t(\bx)$ & Leaf index for data point $\bx$ in tree $t$. \\
$M$ & Dimension of the GBDT embedding. \\
$m$ & Dimension of the transformed feature space after initial transformations (either preprocessing-only, GBDT embedding, or their concatenation). \\
$\bpsi(\bx)$ & Output of the initial transformations.\\
$\sigma(\cdot)$ & pointwise nonlinearity (\eg ReLU).\\
$r$ & Dimension after random feature expansion. \\
$\bOmega$ & Random projection matrix used in the embedding stage. \\
$d_{\text{main}}$ & Dimension of the fixed-size embedding used as input to the main network. \\
$\bU$ & PCA transformation matrix reducing dimension from $r$ to $d_{\text{main}}$. \\
$\bomega$ & Parameters involved in the randomized fixed-size embedding projection.\\
$\Psi_{\bgamma,\bomega}(\cdot)$ & Complete embedding stage function that processes raw features into a fixed-size embedding.\\
$\tilde{\bx}$ & Fixed-size embedding of a data point after the embedding stage. \\
$\btheta$ & Parameters of the main network $f_{\btheta}$. \\
$f_{\btheta}$ & Main network mapping embeddings $\tilde{\bx}$ to output logits in $\R^K$. \\
$f_{\btheta}^{\text{main}}$ & Layers of the main network (feature extractor) except the final classification layer. \\
$f_{\btheta}^{\text{out}}$ & Final classification layer of the main network. \\
$\bphi$ & Parameters of the hypernetwork $g_{\bphi}$ learned during meta-training. \\
$g_{\bphi}$ & Hypernetwork that produces the main network parameters $\btheta$. \\
$\cL$ & Loss function. \\
$\alpha$ & Weight controlling the contribution of retrieval-augmented predictions. \\
$\tau$ & Temperature parameter for scaling similarities in the retrieval mechanism. \\
$\bH$ & Penultimate layer representations from the main network. \\
$\bS$ & Similarity matrix computed between query and context representations. \\
$\bO$ & Output logits. \\
\bottomrule
\end{tabular}
\label{tab:notation}
\end{table}

\begin{figure*}[!t]
    \centering
    \includegraphics[width=\linewidth]{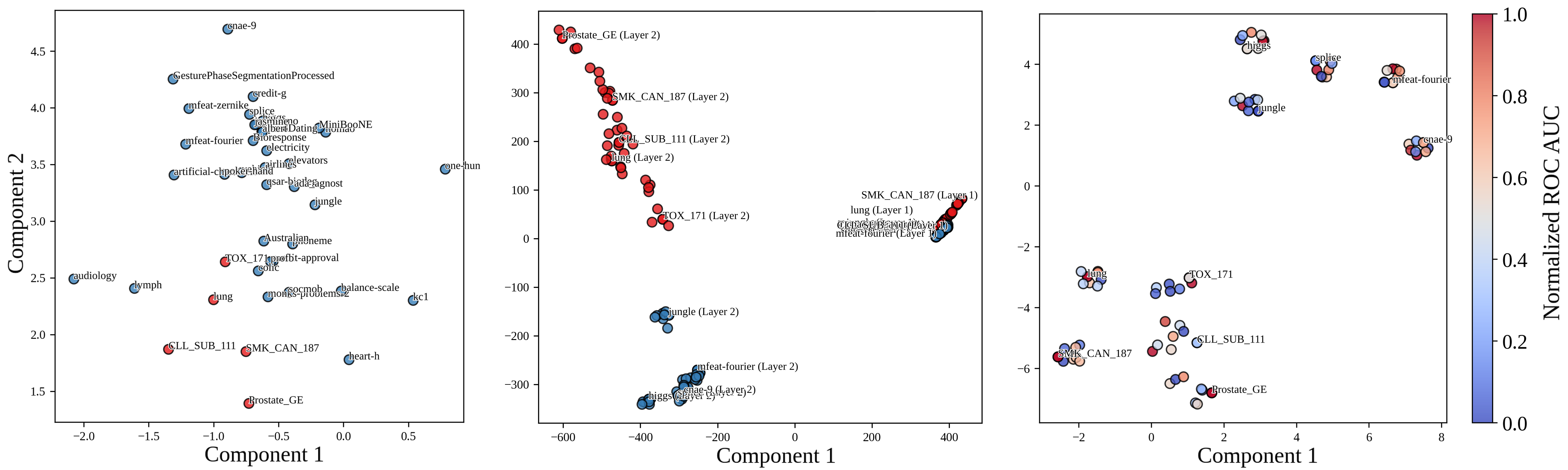}
\caption{Left: t-SNE  of concatenated hypernetwork embeddings across all three layers in embedding space, with blue and red points represent Tabzilla Hard and high dimensional biomedical datasets, respectively. Center: PCA of weights from 8 ensemble predictors trained on each dataset, showing the first two layers separately in weight space. Right: t-SNE  of the weights from the 8 predictors, with weights concatenated across all three layers in weight space, colored by normalized AUC.}
    \label{fig:wsl}
\end{figure*}

\begin{figure*}[h]
    \centering
    \includegraphics[width=1\linewidth]{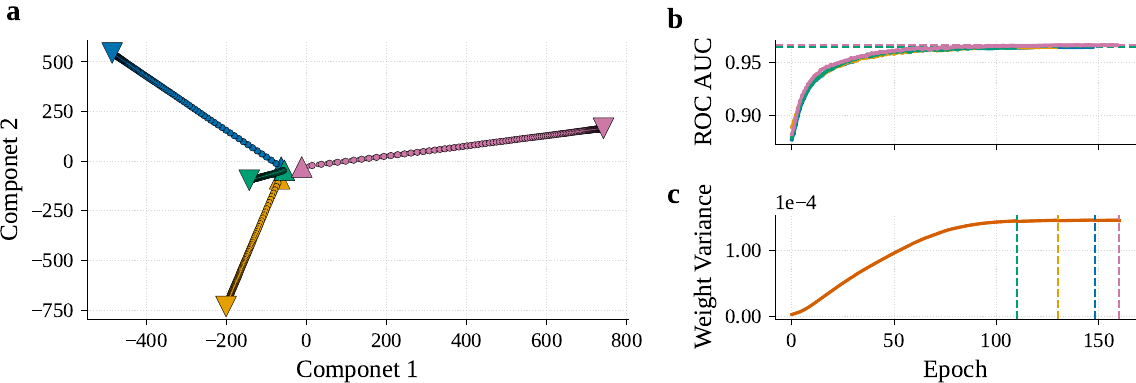}
\caption{Visualization of finetuning on the \emph{jungle-chess} dataset. (a) Evolution of model weights projected onto the first two principal components, with colors for each of four ensemble predictors; triangles indicate start ($\blacktriangle$) and end ($\blacktriangledown$), larger points are epoch boundaries, and smaller points, intermediate iterations. (b) AUC across epochs, with dashed lines showing each predictor’s final performance. (c) Variance of weights across predictors over epochs, with vertical dashed lines marking each predictor’s finetuning completion by early stopping.}
    \label{fig:ftwsl}
\end{figure*}

\section{Further Analysis}


\subsection{Weight Space Analysis}
\label{sec:wsl}

We explore the structure of the weight space induced by our meta-trained hypernetwork and the main networks it produces for different tasks. The weight vectors $\btheta$ produced by the hypernetwork can be regarded as points in a high-dimensional weight space $\Theta$. Recent works \cite{schurholt2022hyper, schurholt2024towards} have studied the weight space of populations of neural networks, providing insights into model characteristics, downstream tasks behavior, and generative modeling. In our framework, the meta-trained hypernetwork learns dataset-level embeddings, which uses to produce populations of neural networks for multiple tasks in a shared weight space.

In \cref{fig:wsl}, we visualize this structure on a representative subset of datasets. 
On the left, we show a t-SNE visualization \cite{van2008tsne} of concatenated \emph{hypernetwork representations} (\ie the dataset-level hyper-representations obtained by the hypernetwork before generating the weights) across all three main layers. 
The embeddings for tasks from the same domain naturally cluster together, indicating that the hypernetwork has learned to generate similar internal representations for datasets sharing certain characteristics. 
Notably, the high-dimensional biomedical tasks (in red) form a cluster that is also close to health-related datasets from TabZilla Hard (in blue), \ie \emph{audiology}, \emph{lymph}, \emph{colic}, and \emph{heart-h}, reflecting domain-level similarities. 
The center plot shows PCA applied to weights of the first two layers produced by eight different predictors generated for each dataset. 
The first layer generated by the hypernetwork shows less variance, as it captures more generic representations across datasets, while the second tends to vary more from one dataset to another, probably learning higher-level representations adapted to each task. 
The right plot depicts the weight vectors colored by normalized ROC AUC per dataset, and concatenated for all layers, which cluster for each dataset.

To better understand how these task-specific weights evolve during fine-tuning, \cref{fig:ftwsl} illustrates trajectories of four ensemble predictors on the \emph{jungle-chess} dataset, part of TabZilla Hard. 
In Fig.~\ref{fig:ftwsl}a, although all predictors begin from closely related initializations generated by the hypernetwork, they diverge over fine-tuning. Displaying distinct trajectories in the two dominant principal components, they gradually become more dissimilar in weight space and thereby increase predictor diversity, a phenomenon beneficial for ensembling, as diversified predictors capture complementary aspects of the data \cite{dietterich2000ensembles,wood2023unified}. 
In Fig. \ref{fig:ftwsl}b, the corresponding ROC AUC increases steadily for each predictor, confirming that the weight trajectories move toward solutions of higher predictive performance. In Fig. \ref{fig:ftwsl}c, the average inter-predictor variance rises as fine-tuning progresses, indicating again that the ensemble members become more diverse.

\subsection{Ablation of Key Hyperparameters}
\label{appendix_ablation_hyperparameters}

The experiments for this ablation study are conducted on the \emph{cylinder-bands}, \emph{ecoli}, \emph{connect-4}, \emph{christine}, and \emph{volkert} datasets from OpenML \cite{OpenML2013, bischl2025openml}, selected for their varied dimensions.
The pre-specified data folds are used for evaluation, and all executions are performed on a single NVIDIA A5500 GPU.

Tables \ref{cls_ablation_auc}, \ref{cls_ablation_fit}, and \ref{cls_ablation_predict} present an ablation study evaluating the impact of key iLTM hyperparameters on AUC, fit time, and prediction time, respectively. 
Namely, they present the performance for a single forward pass on one batch of data (Base), and when adding ensembling of 4 predictors (E), retrieval-augmented classification (R), and fine-tuning of the main network on the full training set (F).

Generally, each additional component tends to improve the AUC score. However, as discussed in the introduction, the efficacy of certain techniques can be dataset-dependent. Consequently, some datasets derive greater benefit from incorporating retrieval or specific embedding strategies (Robust vs. GBDT) than others. Notably, fine-tuning consistently yields the most significant AUC gains.

To identify when retrieval benefits iLTM, we compared retrieval-enabled variants with matched non-retrieval variants on TabZilla and regression benchmarks. Retrieval was not uniformly beneficial: on TabZilla classification tasks, the retrieval variants usually underperformed the non-retrieval ones, winning only 11/36 datasets by AUC. By contrast, retrieval was more effective in regression, where the best retrieval-enabled iLTM variant outperformed the best non-retrieval variant on 23/27 paired datasets. In classification, the strongest gains arose when retrieval was combined with GBDT representations and were positively associated with feature richness, including the number of attributes and the attribute-to-instance ratio. Retrieval often hurt on simple, low-dimensional, or rule-like datasets, including monks-problems-2, balance-scale, poker-hand, jungle-chess, blood-transfusion, and heart-h. These results suggest that retrieval should be validation-selected rather than always enabled, with the greatest expected benefits on regression tasks and feature-rich tabular classification problems.

Regarding fit time, ensembling (x4) leads to an approximately four-fold increase, scaling linearly with the number of predictors. The addition of retrieval incurs a negligible increase in fit time. Fine-tuning, while delivering substantial AUC improvements, also results in the most considerable increase in fit time. Furthermore, the robust embedding requires less time than the GBDT embedding, as it bypasses the need to train a GBDT model and does not expand the input dimensionality.

For prediction time, ensembling (x4) similarly introduces a significant cost, scaling with the number of predictors. In contrast, once an ensemble is in place, the subsequent additions of retrieval and fine-tuning have a relatively minor impact on the overall prediction time.

\begin{table}
    \caption{Ablation study on the AUC for serveral iLTM configurations. Base: single predictor, E: ensembling (4 predictors), R: retrieval-augmented, F: fine-tuning. Results shown for Robust preprocessing and GBDT embeddings.}
    \centering
\begin{tabular}{lcc}

\toprule
\textbf{Configuration} & \textbf{Robust} & \textbf{GBDT} \\
\midrule
Base & $0.859_{\pm0.083}$ & $0.808_{\pm0.043}$ \\
+ E & $0.866_{\pm0.077}$ & $0.838_{\pm0.067}$ \\
+ E + R & $0.855_{\pm0.081}$ & $0.853_{\pm0.091}$ \\
+ E + R + F & $0.927_{\pm0.060}$ & $0.896_{\pm0.073}$ \\
\bottomrule
\end{tabular}
\label{cls_ablation_auc}
\end{table}

\begin{table}
    \caption{Ablation study on the fit time (in seconds) for serveral iLTM configurations. Base: single predictor, E: ensembling (4 predictors), R: retrieval-augmented, F: fine-tuning. Results shown for Robust preprocessing and GBDT embeddings.}
    \centering
\begin{tabular}{lcc}

\toprule
\textbf{Configuration} & \textbf{Robust} & \textbf{GBDT} \\
\midrule
Base & $2.896_{\pm2.308}$ & $7.556_{\pm2.545}$ \\
+ E & $6.404_{\pm3.596}$ & $13.813_{\pm4.467}$ \\
+ E + R & $6.134_{\pm3.763}$ & $13.592_{\pm4.349}$ \\
+ E + R + F & $263.111_{\pm309.644}$ & $313.517_{\pm380.552}$ \\
\bottomrule
\end{tabular}
\label{cls_ablation_fit}
\end{table}

\begin{table}
    \caption{Ablation study on the prediction time (in seconds) for serveral iLTM configurations. Base: single predictor, E: ensembling (4 predictors), R: retrieval-augmented, F: fine-tuning. Results shown for Robust preprocessing and GBDT embeddings.}
    \centering
\begin{tabular}{lcc}

\toprule
\textbf{Configuration} & \textbf{Robust} & \textbf{GBDT} \\
\midrule
Base & $0.719_{\pm0.258}$ & $0.929_{\pm0.246}$ \\
+ E & $2.540_{\pm0.381}$ & $3.367_{\pm0.820}$ \\
+ E + R & $2.563_{\pm0.541}$ & $3.679_{\pm1.204}$ \\
+ E + R + F & $2.519_{\pm0.462}$ & $3.714_{\pm1.155}$ \\
\bottomrule
\end{tabular}
\label{cls_ablation_predict}
\end{table}

\subsection{Controlled Comparison with HyperFast}

To disentangle model design from pretraining scale, we compare iLTM with two HyperFast \cite{bonet2024hyperfast} baselines. We focus on HyperFast because it is the closest prior method to iLTM in its use of real-data meta-training and a hypernetwork that generates task-specific neural-network weights, making it a particularly suitable control for separating the effect of pretraining scale from architectural differences. The published HyperFast model was meta-trained on 83 datasets. We trained a new HyperFast checkpoint, HyperFast-1.8k, on the same 1,806-dataset collection used for iLTM. The original model represents the standard HyperFast pretraining regime, while HyperFast-1.8k matches both the size and composition of iLTM’s pretraining corpus. 

As shown in \cref{hyperfast_vs_iltm}, scaling HyperFast improves its performance, but it remains below iLTM. Therefore, iLTM’s advantage is not explained by pretraining-set size alone and is consistent with benefits from its integrated design.

\begin{table}[!ht]
\caption{Comparison with HyperFast under original and matched-data pretraining. HyperFast-1.8k is trained on the same 1,806-dataset pretraining collection as iLTM.}
\centering
\begin{tabular}{lcc}
\toprule
\textbf{Model} & \textbf{Wins vs. iLTM} & \textbf{Mean AUC} \\
\midrule
HyperFast      & 5/36         & $0.8753_{\pm 0.1132}$   \\
HyperFast-1.8k & 7/36         & $0.8891_{\pm 0.1141}$   \\
iLTM           & ---            & $\mathbf{0.9142}_{\pm 0.0189}$  \\
\bottomrule
\end{tabular}
\label{hyperfast_vs_iltm}
\end{table}

\section{Benchmarks}
\label{appendix_benchmarks}

\subsection{Dataset Characteristics}
\label{appendix_benchmark_datasets}

\subsubsection{TabZilla Hard}

Table \ref{tab:tabhard} shows the number of samples and features included in the 36 datasets of the TabZilla Hard Benchmark Suite \cite{mcelfresh2023tabzilla}.

\begin{table}[h]
    \caption{Dimensions of the TabZilla Hard \cite{mcelfresh2023tabzilla} datasets.}
    \centering
\begin{tabular}{@{}lrr@{}}
\toprule
Dataset               & \# Samples & \# Features \\ \midrule
credit-g              & 1 000      & 21          \\
jungle-chess          & 44 819     & 7           \\
MiniBooNE             & 130 064    & 51          \\
albert                & 425 240    & 79          \\
electricity           & 45 312     & 9           \\
elevators             & 16 599     & 19          \\
guillermo             & 20 000     & 4 297       \\
higgs                 & 98 050     & 29          \\
nomao                 & 34 465     & 119         \\
100-plants-texture    & 1 599      & 65          \\
poker-hand            & 1 025 009  & 11          \\
profb                 & 672        & 10          \\
socmob                & 1 156      & 6           \\
audiology             & 226        & 70          \\
splice                & 3 190      & 61          \\
vehicle               & 846        & 19          \\
Australian            & 690        & 15          \\
Bioresponse           & 3 751      & 1 777       \\
GesturePhase          & 9 872      & 33          \\
SpeedDating           & 8 378      & 121         \\
ada-agnostic          & 4 562      & 49          \\
airlines              & 539 382    & 8           \\
artificial-characters & 10 218     & 8           \\
colic                 & 368        & 27          \\
credit-approval       & 690        & 16          \\
heart-h               & 294        & 14          \\
jasmine               & 2 984      & 145         \\
kc1                   & 2 109      & 22          \\
lymph                 & 148        & 19          \\
mfeat-fourier         & 2 000      & 77          \\
phoneme               & 5 404      & 6           \\
qsar-biodeg           & 1 055      & 42          \\
balance-scale         & 625        & 5           \\
cnae-9                & 1 080      & 857         \\
mfeat-zernike         & 2 000      & 48          \\
monks-problems-2      & 601        & 7           \\ \bottomrule
\end{tabular}
    \label{tab:tabhard}
\end{table}

\subsubsection{High-Dimensional Collection}
\label{appendix_highdim}

While TabZilla Hard includes datasets with varying scales (148 to over 1 million samples) and moderate feature dimensionality (up to 4,297 features), we complement our evaluation with seven additional high-dimensional datasets going up to 19,993 features, primarily from biomedical applications, to demonstrate the scalability of iLTM to higher-dimensional problems, which are common in certain real-world domains. The seven datasets from our curated high-dimensional collection are CLL\_SUB\_11 \cite{Haslinger2004}, Lung \cite{bhattacharjee2001classification}, Prostate-GE \cite{singh2002gene}, SMK\_CAN\_187 \cite{Spira2007}, TOX-171 \cite{bajwa2016cutting}, Arcene \cite{arcene167} and Gisette \cite{gisette170} \footnote{All of them are publicly available online at \url{https://jundongl.github.io/scikit-feature/datasets}}, and their characteristics are listed in Table~\ref{tab:hd_dataset_characteristics}.

\begin{table}[h]
\centering
\caption{High-dimensional datasets characteristics.}
\label{tab:hd_dataset_characteristics}
\begin{tabular}{lrrr}
\toprule
Dataset & \# Samples & \# Features & \# Classes \\
\midrule
CLL\_SUB\_111 & 111 & 11 340 & 3 \\
Lung & 203 & 3 312 & 5 \\
Prostate\_GE & 102 & 5 966 & 2 \\
SMK\_CAN\_187 & 187 & 19 993 & 2 \\
TOX\_171 & 171 & 5 748 & 4 \\
Arcene & 200 & 10 000 & 2 \\
Gisette & 7 000 & 5 000 & 2 \\
\bottomrule
\end{tabular}
\end{table}

\subsubsection{Transfer Learning to Regression Datasets}

For the regression benchmark, we use the publicly available datasets from~\cite{grinsztajn2022tree}, which have been also used in recent studies~\cite{gorishniy2024tabr,gorishniy2025tabm}, described in \cref{tab:regression_dataset_characteristics}.

\begin{table}[h]
\centering
\caption{Dimensions of the regression datasets.}
\label{tab:regression_dataset_characteristics}
\begin{tabular}{lrrr}
\toprule
Dataset & \# Samples & \# Features \\
\midrule
nyc\_taxi\_green\_dec\_2016 & 581\,835 & 9 \\
elevators & 16\,599 & 16 \\
fifa & 18\,063 & 5 \\
wine\_quality & 6\,497 & 11 \\
medical\_charges & 163\,065 & 5 \\
pol & 15\,000 & 26 \\
MiamiHousing2016 & 13\,932 & 14 \\
year & 515\,345 & 90 \\
cpu\_act & 8\,192 & 21 \\
isolet & 7\,797 & 613 \\
Ailerons & 13\,750 & 33 \\
Mercedes\_Benz\_Greener\_Manufacturing & 4\,209 & 359 \\
house\_sales & 21\,613 & 15 \\
particulate\_matter\_ukair\_2017 & 394\,299 & 6 \\
analcatdata\_supreme & 4\,052 & 7 \\
OnlineNewsPopularity & 39\,644 & 59 \\
superconduct & 21\,263 & 79 \\
Brazilian\_houses & 10\,692 & 8 \\
\bottomrule
\end{tabular}
\end{table}

\subsubsection{TabReD Benchmark}

We further evaluate on the TabReD benchmark \cite{rubachev2025tabred}, which consists of eight datasets covering both classification and regression tasks. Table~\ref{tab:tabred} summarizes their characteristics.

\begin{table}[h]
\centering
\caption{Dimensions of the TabReD datasets. For the largest datasets, numbers in parentheses indicate the full dataset size, and we report the subset size  used in TabReD to make hyperparameter tuning feasible, following the original benchmark \cite{rubachev2025tabred}.}
\label{tab:tabred}
\begin{tabular}{lrrl}
\toprule
Dataset & \# Samples & \# Features & Task \\
\midrule
Sberbank Housing & 28\,000 & 392 & Regr. \\
Ecom Offers & 160\,000 & 119 & Class. \\
Homesite Insurance & 260\,000 & 299 & Class. \\
HomeCredit Default & 381\,000 (1.5M) & 696 & Class. \\
Cooking Time & 319\,000 (12.8M) & 192 & Regr. \\
Delivery ETA & 350\,000 (17.0M) & 223 & Regr. \\
Maps Routing & 279\,000 (13.6M) & 986 & Regr. \\
Weather & 423\,000 (16.9M) & 103 & Regr. \\
\bottomrule
\end{tabular}
\end{table}

\subsection{Extended Benchmark Results}

\subsubsection{TabZilla Hard Extended Results}
\label{sec:extended_Results}

\cref{tab:allresults} shows the detailed results for each of the top methods (according to \cref{fig:conover_cd_mean_rank}) across all datasets in the TabZilla Hard Benchmark \cite{mcelfresh2023tabzilla}.  Cells without numerical values available (\texttt{NA}) indicate datasets where the corresponding method failed to run due to out-of-memory errors (\texttt{OOM}), exceeding the time limit (\texttt{TLE}), or runtime or unknown errors (\texttt{ERR}). All errors for methods evaluated in \cite{mcelfresh2023tabzilla} are marked with \texttt{ERR}, as the specific error type distinctions are not provided in the original source. The bottom rows present the average test AUC across all 36 datasets, with missing cells (\texttt{NA}) handled using different computation methods that vary in how they favor methods with incomplete results. Notably, XGBoost \cite{chen2016xgboost} (top 4), TabPFNv2 \cite{hollmann2025accurate} (top 2), and iLTM (top 1), executed successfully across all datasets without errors. Furthermore, iLTM achieved both the best average AUC rank, as demonstrated in \cref{fig:conover_cd_mean_rank}, but also the highest mean test AUC.

\begin{table*}[h]
\centering
\caption{AUC obtained by each of the top models (by average AUC mean rank as shown in Fig. \ref{fig:conover_cd_mean_rank}) in all datasets of the TabZilla Hard Benchmark \cite{mcelfresh2023tabzilla}. For the mean computation at the end, ``NA=0'' indicates that we computed the mean by imputing a 0 for the cells with an error, and ``NA=row avg'' means that we imputed the row average (i.e., average performance across methods for that dataset).}
\label{tab:allresults}
\setlength{\tabcolsep}{4pt}
\begin{tabular}{lccccccccccccccccccccccccccc}
	\toprule
	Dataset \textbackslash\; Method                   & r-MLP           & RMLP            & HF              & RF    & LGBM            & TPFN1           & r-RN            & TabR            & MNCA            & CB              & XGB             & TabM            & TPFN2           & iLTM            \\\midrule
	Australian            & 0.924           & 0.890           & 0.900           & 0.938 & 0.941           & 0.932           & 0.926           & 0.932           & 0.910           & 0.942           & 0.935           & 0.931           & 0.940           & \bfseries 0.943 \\
	Bioresponse           & 0.835           & 0.841           & 0.864           & 0.863 & \bfseries 0.875 & 0.864           & 0.846           & 0.849           & 0.857           & 0.866           & 0.874           & 0.870           & 0.850           & 0.873           \\
	GesturePhase          & 0.728           & 0.892           & 0.829           & 0.854 & 0.902           & 0.829           & 0.787           & \bfseries 0.941 & 0.918           & 0.862           & 0.899           & 0.916           & 0.927           & 0.907           \\
	MiniBooNE             & 0.980           & 0.987           & 0.964           & 0.976 & 0.986           & 0.968           & 0.975           & 0.988           & \ttfamily OOM             & 0.984           & 0.985           & \bfseries 0.989 & 0.980           & 0.988           \\
	SpeedDating           & 0.857           & 0.851           & 0.870           & 0.856 & 0.874           & 0.850           & 0.860           & 0.863           & 0.862           & 0.870           & \bfseries 0.876 & 0.862           & 0.862           & 0.867           \\
	ada-agnostic          & 0.890           & 0.893           & 0.883           & 0.895 & 0.898           & 0.895           & 0.894           & 0.900           & 0.897           & \bfseries 0.906 & 0.902           & 0.899           & 0.904           & 0.901           \\
	airlines              & 0.701           & \ttfamily OOM             & 0.662           & 0.707 & \bfseries 0.725 & 0.624           & 0.705           & \ttfamily OOM             & \ttfamily OOM             & 0.715           & 0.724           & 0.663           & 0.681           & 0.686           \\
	albert                & 0.746           & \ttfamily OOM             & 0.689           & 0.726 & \ttfamily ERR             & 0.689           & 0.760           & \ttfamily OOM             & \ttfamily OOM             & \bfseries 0.777 & 0.759           & \ttfamily OOM             & 0.726           & 0.748           \\
	artificial-characters & 0.916           & 0.986           & 0.962           & 0.971 & 0.996           & 0.961           & 0.954           & 0.996           & 0.997           & 0.988           & \bfseries 0.998 & 0.986           & 0.980           & 0.990           \\
	audiology             & 0.637           & 0.881           & 0.914           & 0.905 & 0.891           & 0.914           & 0.919           & 0.880           & 0.931           & 0.878           & \bfseries 0.938 & 0.902           & 0.932           & 0.914           \\
	balance-scale         & 0.993           & 0.995           & 0.998           & 0.802 & 0.966           & \bfseries 0.999 & 0.996           & 0.994           & 0.995           & 0.950           & 0.918           & 0.994           & 0.995           & 0.996           \\
	cnae-9                & 0.997           & 0.994           & 0.996           & 0.987 & 0.981           & 0.966           & 0.996           & 0.996           & 0.996           & 0.994           & 0.994           & \bfseries 0.998 & 0.997           & 0.998           \\
	colic                 & 0.877           & 0.824           & 0.856           & 0.903 & 0.895           & 0.878           & 0.871           & 0.870           & 0.869           & 0.905           & 0.912           & 0.847           & \bfseries 0.916 & 0.893           \\
	credit-approval       & 0.922           & 0.889           & 0.910           & 0.934 & 0.927           & 0.935           & 0.931           & 0.930           & 0.924           & 0.939           & \bfseries 0.945 & 0.932           & 0.939           & 0.939           \\
	credit-g              & 0.777           & 0.749           & 0.739           & 0.757 & 0.760           & 0.765           & \bfseries 0.796 & 0.769           & 0.758           & 0.775           & 0.774           & 0.771           & 0.787           & 0.785           \\
	electricity           & 0.918           & 0.961           & 0.902           & 0.927 & 0.985           & 0.886           & 0.917           & \bfseries 0.996 & 0.994           & 0.937           & 0.980           & 0.975           & 0.953           & 0.969           \\
	elevators             & 0.760           & 0.952           & 0.949           & 0.894 & 0.940           & 0.945           & 0.794           & 0.951           & 0.953           & 0.944           & 0.944           & \bfseries 0.955 & 0.952           & 0.947           \\
	guillermo             & 0.781           & 0.897           & 0.724           & 0.878 & \ttfamily ERR             & 0.828           & 0.794           & \ttfamily OOM             & \ttfamily OOM             & \ttfamily ERR             & 0.907           & \bfseries 0.913 & 0.845           & 0.897           \\
	heart-h               & \bfseries 0.912 & 0.843           & 0.838           & 0.893 & 0.864           & 0.882           & 0.895           & \ttfamily OOM             & 0.901           & 0.886           & 0.885           & 0.850           & 0.903           & 0.885           \\
	higgs                 & 0.781           & 0.816           & 0.744           & 0.784 & 0.805           & 0.727           & 0.813           & 0.815           & \ttfamily OOM             & 0.802           & 0.805           & \bfseries 0.825 & 0.794           & 0.806           \\
	jasmine               & 0.851           & 0.854           & 0.858           & 0.870 & 0.864           & 0.869           & 0.854           & 0.875           & 0.864           & 0.869           & 0.869           & 0.873           & \bfseries 0.887 & 0.873           \\
	jungle-chess          & 0.967           & \bfseries 1.000 & 0.943           & 0.954 & 0.976           & 0.934           & 0.968           & 1.000           & 0.999           & 0.970           & 0.974           & 1.000           & 0.971           & 0.980           \\
	kc1                   & 0.781           & 0.791           & 0.807           & 0.820 & 0.793           & 0.828           & 0.796           & 0.817           & 0.810           & 0.815           & 0.807           & 0.813           & 0.829           & \bfseries 0.835 \\
	lymph                 & 0.852           & 0.856           & 0.877           & 0.903 & 0.851           & 0.909           & 0.909           & 0.851           & 0.897           & 0.886           & 0.925           & 0.927           & \bfseries 0.927 & 0.912           \\
	mfeat-fourier         & 0.973           & 0.984           & 0.982           & 0.980 & 0.981           & 0.988           & 0.978           & 0.980           & 0.983           & 0.983           & 0.983           & 0.987           & \bfseries 0.991 & 0.985           \\
	mfeat-zernike         & 0.979           & 0.983           & 0.980           & 0.972 & 0.974           & 0.981           & 0.981           & \bfseries 0.989 & 0.981           & 0.976           & 0.973           & 0.982           & 0.988           & 0.984           \\
	monks-problems-2      & \bfseries 1.000 & \bfseries 1.000 & \bfseries 1.000 & 0.976 & 0.988           & \bfseries 1.000 & \bfseries 1.000 & 0.994           & \bfseries 1.000 & 0.981           & 0.999           & \bfseries 1.000 & \bfseries 1.000 & 0.999           \\
	nomao                 & 0.992           & 0.993           & 0.991           & 0.992 & \bfseries 0.996 & 0.988           & 0.993           & 0.994           & 0.995           & 0.995           & 0.996           & 0.996           & 0.993           & 0.995           \\
	100-plants-texture    & 0.891           & 0.996           & \ttfamily ERR             & 0.988 & \ttfamily ERR             & 0.996           & 0.992           & 0.997           & 0.997           & 0.997           & 0.992           & \bfseries 0.998 & 0.997           & 0.997           \\
	phoneme               & 0.948           & 0.956           & 0.942           & 0.955 & 0.959           & 0.942           & 0.936           & 0.964           & \bfseries 0.968 & 0.945           & 0.960           & 0.962           & 0.968           & 0.966           \\
	poker-hand            & 0.546           & \ttfamily OOM             & 0.589           & \ttfamily ERR   & \ttfamily ERR             & 0.527           & \ttfamily ERR             & \ttfamily OOM             & \ttfamily OOM             & 0.934           & 0.860           & \bfseries 0.995 & 0.671           & 0.803           \\
	profb                 & 0.727           & 0.567           & 0.629           & 0.640 & 0.694           & 0.662           & 0.703           & 0.561           & 0.733           & \bfseries 0.773 & 0.664           & 0.584           & 0.663           & 0.758           \\
	qsar-biodeg           & 0.923           & 0.922           & 0.936           & 0.926 & 0.918           & \bfseries 0.941 & 0.927           & 0.921           & 0.898           & 0.930           & 0.928           & 0.936           & 0.939           & 0.937           \\
	socmob                & 0.984           & 0.968           & 0.975           & 0.977 & 0.977           & 0.978           & 0.985           & 0.989           & 0.967           & 0.986           & 0.988           & 0.989           & \bfseries 0.990 & 0.985           \\
	splice                & 0.990           & \ttfamily ERR             & 0.966           & 0.992 & 0.993           & 0.978           & 0.991           & 0.959           & 0.993           & 0.992           & 0.993           & 0.990           & \bfseries 0.995 & 0.994           \\
	vehicle               & 0.830           & 0.946           & 0.967           & 0.927 & 0.920           & 0.969           & 0.952           & 0.967           & 0.966           & 0.934           & 0.934           & 0.956           & \bfseries 0.973 & 0.960           \\\midrule
	Mean (\texttt{NA}=0)           & 0.866           & 0.804           & 0.851           & 0.870 & 0.808           & 0.884           & 0.872           & 0.792           & 0.773           & 0.886           & 0.911           & 0.891           & 0.907           & \bfseries 0.914 \\
	Mean (\texttt{NA}=row avg)     & 0.866           & 0.892           & 0.878           & 0.891 & 0.900           & 0.884           & 0.893           & 0.901           & 0.906           & 0.909           & 0.911           & 0.911           & 0.907           & \bfseries 0.914 \\
    \bottomrule
\end{tabular}
\end{table*}

We note that although here we provide the mean test AUC for completeness, average AUC can be biased by scale differences and missing data, while average ranking provides a more robust measure of relative performance across diverse datasets and supports rigorous statistical testing via the Nemenyi test \cite{demvsar2006statistical}. Critical difference diagrams with average rankings, as we have shown in \cref{fig:conover_cd_mean_rank} are becoming the standard evaluation approach in tabular ML \cite{mcelfresh2023tabzilla,ye2025modernnca,gorishniy2025tabm}.

\cref{tab:algorithm_scores} presents the detailed AUC scores with standard deviations computed over the cross-validation folds for XGBoost \cite{chen2016xgboost}, TabPFNv2 \cite{hollmann2025accurate}, and iLTM on the TabZilla Hard Benchmark \cite{mcelfresh2023tabzilla}. While all three methods exhibit comparable standard deviations, iLTM shows slightly higher variability and TabPFNv2 demonstrates slightly lower variability across the cross-validation folds.

\begin{table*}[!ht]
\centering
\caption{Detailed AUC scores with standard deviations for XGBoost \cite{chen2016xgboost}, TabPFNv2 \cite{hollmann2025accurate}, and iLTM on the TabZilla Hard Benchmark \cite{mcelfresh2023tabzilla}.}
\begin{tabular}{lccccc}
\toprule
Dataset & XGBoost & TabPFNv2 & iLTM \\
\midrule
Australian & $0.9346_{\pm 0.0209}$ & $0.9398_{\pm 0.0251}$ & $0.9434_{\pm 0.0215}$ \\
Bioresponse & $0.8741_{\pm 0.0160}$ & $0.8496_{\pm 0.0247}$ & $0.8729_{\pm 0.0174}$ \\
GesturePhase & $0.8986_{\pm 0.0095}$ & $0.9270_{\pm 0.0060}$ & $0.9067_{\pm 0.0085}$ \\
MiniBooNE & $0.9853_{\pm 0.0011}$ & $0.9797_{\pm 0.0017}$ & $0.9884_{\pm 0.0010}$ \\
SpeedDating & $0.8764_{\pm 0.0114}$ & $0.8617_{\pm 0.0119}$ & $0.8665_{\pm 0.0126}$ \\
ada-agnostic & $0.9019_{\pm 0.0187}$ & $0.9043_{\pm 0.0144}$ & $0.9010_{\pm 0.0174}$ \\
airlines & $0.7242_{\pm 0.0024}$ & $0.6812_{\pm 0.0026}$ & $0.6902_{\pm 0.0029}$ \\
albert & $0.7586_{\pm 0.0027}$ & $0.7256_{\pm 0.0018}$ & $0.7592_{\pm 0.0024}$ \\
artificial-characters & $0.9980_{\pm 0.0004}$ & $0.9804_{\pm 0.0015}$ & $0.9903_{\pm 0.0012}$ \\
audiology & $0.9382_{\pm 0.0479}$ & $0.9318_{\pm 0.0409}$ & $0.9140_{\pm 0.0553}$ \\
balance-scale & $0.9178_{\pm 0.0332}$ & $0.9954_{\pm 0.0044}$ & $0.9955_{\pm 0.0050}$ \\
cnae-9 & $0.9942_{\pm 0.0039}$ & $0.9972_{\pm 0.0042}$ & $0.9981_{\pm 0.0018}$ \\
colic & $0.9116_{\pm 0.0480}$ & $0.9163_{\pm 0.0470}$ & $0.8926_{\pm 0.0603}$ \\
credit-approval & $0.9446_{\pm 0.0391}$ & $0.9394_{\pm 0.0371}$ & $0.9387_{\pm 0.0434}$ \\
credit-g & $0.7742_{\pm 0.0390}$ & $0.7868_{\pm 0.0413}$ & $0.7845_{\pm 0.0444}$ \\
electricity & $0.9804_{\pm 0.0013}$ & $0.9533_{\pm 0.0025}$ & $0.9692_{\pm 0.0019}$ \\
elevators & $0.9437_{\pm 0.0052}$ & $0.9524_{\pm 0.0051}$ & $0.9469_{\pm 0.0054}$ \\
guillermo & $0.9073_{\pm 0.0080}$ & $0.8455_{\pm 0.0159}$ & $0.8969_{\pm 0.0106}$ \\
heart-h & $0.8849_{\pm 0.0439}$ & $0.9025_{\pm 0.0374}$ & $0.8853_{\pm 0.0438}$ \\
higgs & $0.8051_{\pm 0.0072}$ & $0.7938_{\pm 0.0074}$ & $0.8063_{\pm 0.0067}$ \\
jasmine & $0.8694_{\pm 0.0181}$ & $0.8873_{\pm 0.0122}$ & $0.8734_{\pm 0.0130}$ \\
jungle-chess & $0.9742_{\pm 0.0015}$ & $0.9707_{\pm 0.0059}$ & $0.9799_{\pm 0.0013}$ \\
kc1 & $0.8073_{\pm 0.0480}$ & $0.8293_{\pm 0.0290}$ & $0.8347_{\pm 0.0364}$ \\
lymph & $0.9252_{\pm 0.0415}$ & $0.9267_{\pm 0.0645}$ & $0.9120_{\pm 0.1028}$ \\
mfeat-fourier & $0.9831_{\pm 0.0032}$ & $0.9914_{\pm 0.0022}$ & $0.9854_{\pm 0.0028}$ \\
mfeat-zernike & $0.9729_{\pm 0.0052}$ & $0.9878_{\pm 0.0030}$ & $0.9837_{\pm 0.0030}$ \\
monks-problems-2 & $0.9985_{\pm 0.0020}$ & $1.0000_{\pm 0.0000}$ & $0.9992_{\pm 0.0014}$ \\
nomao & $0.9957_{\pm 0.0010}$ & $0.9928_{\pm 0.0008}$ & $0.9951_{\pm 0.0007}$ \\
100-plants-texture & $0.9925_{\pm 0.0030}$ & $0.9966_{\pm 0.0013}$ & $0.9969_{\pm 0.0014}$ \\
phoneme & $0.9602_{\pm 0.0093}$ & $0.9678_{\pm 0.0081}$ & $0.9662_{\pm 0.0106}$ \\
poker-hand & $0.8602_{\pm 0.0385}$ & $0.6707_{\pm 0.0141}$ & $0.8025_{\pm 0.0207}$ \\
profb & $0.6642_{\pm 0.0620}$ & $0.6628_{\pm 0.0547}$ & $0.7577_{\pm 0.0722}$ \\
qsar-biodeg & $0.9283_{\pm 0.0312}$ & $0.9393_{\pm 0.0298}$ & $0.9366_{\pm 0.0301}$ \\
socmob & $0.9879_{\pm 0.0095}$ & $0.9900_{\pm 0.0079}$ & $0.9854_{\pm 0.0063}$ \\
splice & $0.9933_{\pm 0.0025}$ & $0.9952_{\pm 0.0030}$ & $0.9945_{\pm 0.0026}$ \\
vehicle & $0.9344_{\pm 0.0102}$ & $0.9734_{\pm 0.0073}$ & $0.9603_{\pm 0.0104}$ \\
\midrule
Average & $0.9111_{\pm 0.0180}$ & $0.9068_{\pm 0.0160}$ & $\mathbf{0.9142}_{\pm 0.0189}$ \\
\bottomrule
\end{tabular}
\label{tab:algorithm_scores}
\end{table*}

\subsubsection{Extended Regression Results} 
Following the tuning and evaluation procedure used in~\cite{gorishniy2024tabr,gorishniy2025tabm}, each model is evaluated on each dataset under 15 random seeds. In \cref{tab:test_rmse} we report the mean test root mean square error (RMSE) and its standard deviation over the random seeds. 

\begin{table*}[h]
\caption{Test RMSE ($\downarrow$) scores across regression datasets.}
\label{tab:test_rmse}
\centering
\scriptsize
\setlength{\tabcolsep}{2pt}
\renewcommand{\arraystretch}{0.95}
\resizebox{\textwidth}{!}{%
\begin{tabular}{p{3cm}rrrrrrrrrr}
\toprule
Dataset 
        & \multicolumn{1}{c}{MLP}
        & \multicolumn{1}{c}{FT-Transformer}
        & \multicolumn{1}{c}{MLP-PLR}
        & \multicolumn{1}{c}{LightGBM}
        & \multicolumn{1}{c}{XGBoost}
        & \multicolumn{1}{c}{TabR}
        & \multicolumn{1}{c}{TabM}
        & \multicolumn{1}{c}{CatBoost}
        & \multicolumn{1}{c}{TabPFNv2}
        & \multicolumn{1}{c}{iLTM} \\
\midrule
Nyc-Taxi. 2016 & $0.3966_{\pm 0.001}$ & $0.3907_{\pm 0.001}$ & $0.3684_{\pm 0.002}$ & $0.3688_{\pm 0.000}$ & $0.3792_{\pm 0.000}$ & $0.3725_{\pm 0.009}$ & $0.3849_{\pm 0.001}$ & $0.3647_{\pm 0.001}$ & $0.4044_{\pm 0.006}$ & $0.3605_{\pm 0.000}$ \\
Elevators & $0.0051_{\pm 0.000}$ & $0.0050_{\pm 0.000}$ & $0.0024_{\pm 0.000}$ & $0.0020_{\pm 0.000}$ & $0.0020_{\pm 0.000}$ & $0.0019_{\pm 0.000}$ & $0.0019_{\pm 0.000}$ & $0.0020_{\pm 0.000}$ & $0.0019_{\pm 0.000}$ & $0.0018_{\pm 0.000}$ \\
Fifa & $0.8025_{\pm 0.001}$ & $0.7929_{\pm 0.002}$ & $0.7945_{\pm 0.002}$ & $0.7807_{\pm 0.001}$ & $0.7799_{\pm 0.001}$ & $0.7914_{\pm 0.003}$ & $0.7953_{\pm 0.002}$ & $0.7835_{\pm 0.001}$ & $0.7821_{\pm 0.001}$ & $0.7796_{\pm 0.001}$ \\
Wine Quality & $0.6716_{\pm 0.006}$ & $0.6762_{\pm 0.004}$ & $0.6543_{\pm 0.006}$ & $0.6135_{\pm 0.002}$ & $0.6039_{\pm 0.002}$ & $0.6412_{\pm 0.008}$ & $0.6328_{\pm 0.004}$ & $0.6088_{\pm 0.002}$ & $0.6918_{\pm 0.002}$ & $0.6055_{\pm 0.003}$ \\
Medical Charges & $0.0816_{\pm 0.000}$ & $0.0814_{\pm 0.000}$ & $0.0811_{\pm 0.000}$ & $0.0820_{\pm 0.000}$ & $0.0825_{\pm 0.000}$ & $0.0811_{\pm 0.000}$ & $0.0812_{\pm 0.000}$ & $0.0816_{\pm 0.000}$ & $0.0813_{\pm 0.000}$ & $0.0813_{\pm 0.000}$ \\
Pol & $5.6589_{\pm 0.206}$ & $2.7528_{\pm 0.133}$ & $2.6145_{\pm 0.127}$ & $4.2320_{\pm 0.036}$ & $4.2964_{\pm 0.022}$ & $2.5770_{\pm 0.159}$ & $3.0198_{\pm 0.061}$ & $3.6319_{\pm 0.041}$ & $3.3983_{\pm 0.107}$ & $2.8480_{\pm 0.032}$ \\
Miamihousing2016 & $0.1614_{\pm 0.002}$ & $0.1517_{\pm 0.002}$ & $0.1502_{\pm 0.001}$ & $0.1461_{\pm 0.000}$ & $0.1440_{\pm 0.000}$ & $0.1392_{\pm 0.001}$ & $0.1477_{\pm 0.001}$ & $0.1417_{\pm 0.001}$ & $0.1348_{\pm 0.000}$ & $0.1386_{\pm 0.000}$ \\
Year & $8.9635_{\pm 0.018}$ & $9.0127_{\pm 0.020}$ & $8.9274_{\pm 0.013}$ & $9.0200_{\pm 0.003}$ & $9.0307_{\pm 0.003}$ & $8.9721_{\pm 0.011}$ & $8.8705_{\pm 0.004}$ & $9.0370_{\pm 0.007}$ & $9.0328_{\pm 0.022}$ & $8.8884_{\pm 0.010}$ \\
Cpu Act & $2.7119_{\pm 0.175}$ & $2.2356_{\pm 0.081}$ & $2.2700_{\pm 0.029}$ & $2.2223_{\pm 0.015}$ & $2.5237_{\pm 0.068}$ & $2.1278_{\pm 0.052}$ & $2.1402_{\pm 0.011}$ & $2.1239_{\pm 0.015}$ & $2.6881_{\pm 0.016}$ & $2.1233_{\pm 0.015}$ \\
Isolet & $2.2226_{\pm 0.143}$ & $2.3317_{\pm 0.163}$ & $2.2240_{\pm 0.106}$ & $2.7005_{\pm 0.013}$ & $2.7567_{\pm 0.017}$ & $1.9919_{\pm 0.134}$ & $1.8433_{\pm 0.039}$ & $2.8671_{\pm 0.014}$ & $1.9657_{\pm 0.029}$ & $1.9787_{\pm 0.032}$ \\
Ailerons & $0.0002_{\pm 0.000}$ & $0.0002_{\pm 0.000}$ & $0.0002_{\pm 0.000}$ & $0.0002_{\pm 0.000}$ & $0.0002_{\pm 0.000}$ & $0.0002_{\pm 0.000}$ & $0.0002_{\pm 0.000}$ & $0.0002_{\pm 0.000}$ & $0.0002_{\pm 0.000}$ & $0.0002_{\pm 0.000}$ \\
Mercedes Benz G. M. & $8.3828_{\pm 0.060}$ & $8.2517_{\pm 0.068}$ & $8.3828_{\pm 0.060}$ & $8.2078_{\pm 0.043}$ & $8.2177_{\pm 0.019}$ & $8.3187_{\pm 0.075}$ & $8.2052_{\pm 0.026}$ & $8.1629_{\pm 0.022}$ & $8.1288_{\pm 0.008}$ & $8.2155_{\pm 0.026}$ \\
House Sales & $0.1813_{\pm 0.001}$ & $0.1687_{\pm 0.001}$ & $0.1687_{\pm 0.001}$ & $0.1692_{\pm 0.000}$ & $0.1694_{\pm 0.000}$ & $0.1636_{\pm 0.001}$ & $0.1666_{\pm 0.000}$ & $0.1669_{\pm 0.000}$ & $0.1599_{\pm 0.000}$ & $0.1648_{\pm 0.000}$ \\
Particulate-Mattr & $0.3766_{\pm 0.001}$ & $0.3734_{\pm 0.001}$ & $0.3670_{\pm 0.001}$ & $0.3637_{\pm 0.000}$ & $0.3641_{\pm 0.000}$ & $0.3596_{\pm 0.000}$ & $0.3671_{\pm 0.001}$ & $0.3647_{\pm 0.000}$ & $0.3761_{\pm 0.001}$ & $0.3641_{\pm 0.001}$ \\
Analcatdata Supreme & $0.0777_{\pm 0.002}$ & $0.0781_{\pm 0.001}$ & $0.0789_{\pm 0.003}$ & $0.0778_{\pm 0.002}$ & $0.0801_{\pm 0.003}$ & $0.0807_{\pm 0.007}$ & $0.0786_{\pm 0.002}$ & $0.0780_{\pm 0.006}$ & $0.0736_{\pm 0.001}$ & $0.0817_{\pm 0.001}$ \\
Onlinenewspopularity & $0.8620_{\pm 0.001}$ & $0.8639_{\pm 0.002}$ & $0.8618_{\pm 0.001}$ & $0.8546_{\pm 0.000}$ & $0.8545_{\pm 0.000}$ & $0.8624_{\pm 0.001}$ & $0.8579_{\pm 0.000}$ & $0.8532_{\pm 0.000}$ & $0.8532_{\pm 0.000}$ & $0.8563_{\pm 0.000}$ \\
Superconduct & $10.7241_{\pm 0.062}$ & $10.7765_{\pm 0.111}$ & $10.5663_{\pm 0.058}$ & $10.1634_{\pm 0.012}$ & $10.1610_{\pm 0.020}$ & $10.3835_{\pm 0.056}$ & $10.2628_{\pm 0.028}$ & $10.2422_{\pm 0.022}$ & $9.9964_{\pm 0.025}$ & $10.3090_{\pm 0.020}$ \\
Brazilian Houses & $0.0487_{\pm 0.003}$ & $0.0457_{\pm 0.004}$ & $0.0428_{\pm 0.003}$ & $0.0603_{\pm 0.001}$ & $0.0541_{\pm 0.001}$ & $0.0451_{\pm 0.005}$ & $0.0418_{\pm 0.001}$ & $0.0468_{\pm 0.002}$ & $0.0194_{\pm 0.001}$ & $0.0534_{\pm 0.001}$ \\
\midrule
Avg. Rank & $5.8 \pm 1.9$ & $5.0 \pm 2.0$ & $4.6 \pm 1.8 $ & $4.1 \pm 1.8$ & $4.0 \pm 1.8$ & $3.7 \pm 1.9$ & $3.4 \pm 1.7$ & $3.0 \pm 1.4$ & $2.9 \pm 2.2$ & $\mathbf{2.7} \pm 1.6$ \\
\bottomrule
\end{tabular}
}
\end{table*}

\subsubsection{Detailed TabReD Results}
\label{appendix:tabred_results}

\cref{tab:tabular_benchmarks} shows the complete results for the TabReD benchmark.

\begin{table*}[h]
\centering
\caption{TabReD classification and regression benchmarks (mean $\pm$ std) performance comparison.}
\label{tab:tabular_benchmarks}
\setlength{\tabcolsep}{4.5pt}
\begin{tabular}{lccc ccccc c}
\toprule
\multirow{3}{*}{Methods} & \multicolumn{3}{c}{Classification (ROC AUC $\uparrow$)} & \multicolumn{5}{c}{Regression (RMSE $\downarrow$)} & \multirow{3}{*}{\shortstack{Average \\ Rank $\downarrow$}} \\
\cmidrule(lr){2-4}\cmidrule(lr){5-9}
& \shortstack{Homesite \\ Insurance} 
& \shortstack{Ecom \\ Offers} 
& \shortstack{HomeCredit \\ Default} 
& \shortstack{Sberbank \\ Housing} 
& \shortstack{Cooking \\ Time} 
& \shortstack{Delivery \\ ETA} 
& \shortstack{Maps \\ Routing} 
& \shortstack{Weather \vspace{1.2ex}} 
& \\
\midrule

\multicolumn{10}{l}{\textbf{Non DL Baselines}}\\
XGBoost \cite{chen2016xgboost}        & $0.960_{\pm0.000}$ & $0.576_{\pm0.007}$ & $0.867_{\pm0.001}$ & $0.242_{\pm0.001}$ & $0.482_{\pm0.000}$ & $0.547_{\pm0.000}$ & $0.162_{\pm0.000}$ & $1.467_{\pm0.001}$ & $2.9_{\pm1.6}$ \\
LightGBM \cite{ke2017lightgbm}       & $0.960_{\pm0.000}$ & $0.576_{\pm0.001}$ & $0.866_{\pm0.000}$ & $0.247_{\pm0.001}$ & $0.483_{\pm0.000}$ & $0.547_{\pm0.000}$ & $0.162_{\pm0.000}$ & $1.462_{\pm0.001}$ & $3.3_{\pm1.6}$ \\
CatBoost \cite{prokhorenkova2018catboost}       & $0.961_{\pm0.000}$ & $0.560_{\pm0.007}$ & $0.862_{\pm0.001}$ & $0.248_{\pm0.003}$ & $0.482_{\pm0.000}$ & $0.546_{\pm0.000}$ & $0.162_{\pm0.000}$ & $1.469_{\pm0.002}$ & $3.6_{\pm1.4}$ \\
\midrule

\multicolumn{10}{l}{\textbf{Tabular DL Models}}\\
MLP \cite{gorishniy2021revisiting}           & $0.950_{\pm0.000}$ & $0.602_{\pm0.001}$ & $0.855_{\pm0.001}$ & $0.251_{\pm0.005}$ & $0.482_{\pm0.000}$ & $0.550_{\pm0.001}$ & $0.162_{\pm0.000}$ & $1.547_{\pm0.004}$ & $4.6\pm2.1$ \\
SNN \cite{klambauer2017self}           & $0.949_{\pm0.002}$ & $0.600_{\pm0.003}$ & $0.855_{\pm0.001}$ & $0.286_{\pm0.045}$ & $0.484_{\pm0.001}$ & $0.554_{\pm0.003}$ & $0.165_{\pm0.001}$ & $1.565_{\pm0.009}$ & $6.3\pm2.1$ \\
DFR \cite{kirichenko2023last}            & $0.950_{\pm0.000}$ & $0.601_{\pm0.001}$ & $0.855_{\pm0.001}$ & $0.249_{\pm0.004}$ & $0.482_{\pm0.000}$ & $0.551_{\pm0.000}$ & $0.163_{\pm0.000}$ & $1.551_{\pm0.004}$ & $5.0\pm2.3$ \\
DCNv2 \cite{wang2021dcn}          & $0.939_{\pm0.006}$ & $0.596_{\pm0.007}$ & $0.847_{\pm0.003}$ & $0.277_{\pm0.016}$ & $0.484_{\pm0.001}$ & $0.553_{\pm0.002}$ & $0.167_{\pm0.001}$ & $1.578_{\pm0.006}$ & $7.1\pm2.5$ \\
ResNet \cite{gorishniy2021revisiting}         & $0.947_{\pm0.002}$ & $0.600_{\pm0.004}$ & $0.849_{\pm0.003}$ & $0.274_{\pm0.034}$ & $0.482_{\pm0.000}$ & $0.553_{\pm0.002}$ & $0.163_{\pm0.000}$ & $1.502_{\pm0.004}$ & $5.5\pm1.9$ \\
FT-Transformer \cite{gorishniy2021revisiting} & $0.962_{\pm0.001}$ & $0.578_{\pm0.006}$ & $0.857_{\pm0.002}$ & $0.244_{\pm0.004}$ & $0.482_{\pm0.001}$ & $0.554_{\pm0.003}$ & $0.163_{\pm0.000}$ & $1.510_{\pm0.010}$ & $4.6\pm1.7$ \\
MLP-PLR \cite{gorishniy2022embeddings}        & $0.962_{\pm0.000}$ & $0.596_{\pm0.003}$ & $0.857_{\pm0.001}$ & $0.244_{\pm0.005}$ & $0.481_{\pm0.001}$ & $0.553_{\pm0.002}$ & $0.162_{\pm0.000}$ & $1.518_{\pm0.006}$ & $3.6\pm1.8$ \\
TabR \cite{gorishniy2024tabr}           & $0.949_{\pm0.001}$ & $0.594_{\pm0.002}$ & $0.850_{\pm0.003}$ & $0.282_{\pm0.032}$ & $0.483_{\pm0.001}$ & $0.551_{\pm0.002}$ & $0.164_{\pm0.000}$ & $1.467_{\pm0.004}$ & $5.4\pm2.1$ \\
TabM \cite{gorishniy2025tabm}           & $0.964_{\pm0.000}$ & $0.594_{\pm0.000}$ & $0.860_{\pm0.001}$ & $0.244_{\pm0.002}$ & $0.480_{\pm0.000}$ & $0.549_{\pm0.000}$ & $0.161_{\pm0.000}$ & $1.472_{\pm0.002}$ & $2.6\pm1.5$ \\
\midrule
\textbf{iLTM (Ours)}           & $0.961_{\pm0.000}$ & $0.592_{\pm0.002}$ & $0.865_{\pm0.001}$ & $0.237_{\pm0.001}$ & $0.482_{\pm0.001}$ & $0.548_{\pm0.002}$ & $0.162_{\pm0.000}$ & $1.464_{\pm0.004}$ & $\mathbf{2.5}\pm\mathbf{0.9}$ \\
\bottomrule
\end{tabular}
\end{table*}

\subsection{Hyperparameter Search Spaces}

\label{appendix_hparams}

\paragraph{iLTM}

The hyperparameter search space defined for iLTM is shown in Table~\ref{tab:hyperparams}. Note that fine-tuning and retrieval specific hyperparameters only have an effect if ``Do finetuning'' and ``Do retrieval'' are True, respectively. Similarly, GBDT embedding hyperparameters only have an effect if the preprocessing uses them, that is, if the preprocessing is not R.
\begin{table*}[htbp]
\centering
\caption{iLTM Hyperparameter Search Space}
\label{tab:hyperparams}
\begin{tabular}{llll}
\toprule
\textbf{Hyperparameter} & \textbf{Type} & \textbf{Range} & \textbf{Default} \\
\midrule
\multicolumn{4}{l}{\textit{General}} \\
Preprocessing & Categorical & \{R, X, C, RX, RC\} & RX \\
Batch size & Categorical & \{1024, 2048\} & 2048 \\
Number of ensembles & Categorical & \{1, 2, 4, 8, 12, 16, 20\} & 8 \\
Feature bagging & Categorical & \{True, False\} & True \\
\midrule
\multicolumn{4}{l}{\textit{Finetuning}} \\
Do finetuning & Categorical & \{True, False\} & True \\
Dropout & Categorical & \{0.0, 0.15\} & 0.0 \\
Max steps & Categorical & \{4, 512, 1024\} & 1024 \\
Learning rate & Log-uniform & [$10^{-6}$, $10^{-2}$] & $10^{-4}$ \\
Data & Categorical & \{Bootstrap, Entire dataset\} & Entire dataset \\
\midrule
\multicolumn{4}{l}{\textit{GBDT Embedding}} \\
Data split & Categorical & \{Dynamic, Entire dataset\} & Dynamic \\
Fit for each predictor & Categorical & \{True, False\} & False \\
Number of estimators & Categorical & \{100, 300\} & 100 \\
Learning rate & Log-uniform & [0.01, 0.5] & GBDT model default \\
\midrule
\multicolumn{4}{l}{\textit{Retrieval}} \\
Do retrieval & Categorical & \{True, False\} & True \\
Temperature $\tau$ & Uniform & [0.5, 3] & 2.0 \\
Weight $\alpha$ & Uniform & [0, 1] & 0.5 \\
\bottomrule
\end{tabular}
\end{table*}

\paragraph{LogReg-new}

\cref{tab:logreg_params} shows the hyperparameter search space defined for our re-execution of logistic regression. 

\begin{table*}[htbp]
\centering
\caption{LogReg-new Hyperparameter Search Space}
\label{tab:logreg_params}
\begin{tabular}{llll}
\toprule
\textbf{Hyperparameter} & \textbf{Type} & \textbf{Range} & \textbf{Default} \\
\midrule
Regularization (C) & Log-uniform & [$10^{-5}$, $10^3$] & 1.0 \\
Penalty & Categorical & \{L1, L2\} & L2 \\
\bottomrule
\end{tabular}
\end{table*}

\paragraph{HyperFast}

The hyperparameter search space for evaluating HyperFast \cite{bonet2024hyperfast} on the TabZilla Hard Benchmark \cite{mcelfresh2023tabzilla} is presented in \cref{tab:hyperfast_params}. These hyperparameter ranges were selected based on the recommendations provided in the official HyperFast repository documentation on GitHub\footnote{https://github.com/AI-sandbox/HyperFast}.

\begin{table*}[htbp]
\centering
\caption{HyperFast Hyperparameter Search Space}
\label{tab:hyperfast_params}
\begin{tabular}{llll}
\toprule
\textbf{Hyperparameter} & \textbf{Type} & \textbf{Range} & \textbf{Default} \\
\midrule
\multicolumn{4}{l}{\textit{General}} \\
Number of ensembles & Categorical & \{1, 4, 8, 16, 32\} & Model default \\
Batch size & Categorical & \{1024, 2048\} & Model default  \\
Neural network bias & Categorical & \{True, False\} & Model default  \\
Stratified sampling & Categorical & \{True, False\} & Model default  \\
\midrule
\multicolumn{4}{l}{\textit{Optimization}} \\
Optimization type & Categorical & \{None, optimize, ensemble\_optimize\} & Model default  \\
Optimization steps & Categorical & \{1, 4, 8, 16, 32, 64, 128\} & Model default  \\
Random seed & Categorical & \{0, 1, 2, 3, 4, 5, 6, 7, 8, 9\} & Model default  \\
\bottomrule
\end{tabular}
\end{table*}

\paragraph{TabR}

The hyperparameter search space for TabR \cite{gorishniy2024tabr} on the TabZilla Hard Benchmark \cite{mcelfresh2023tabzilla} is presented in \cref{tab:tabr_hyperparams}. These hyperparameter ranges were selected based on the recommendations provided in the paper.

\begin{table*}[htbp]
\centering
\caption{TabR Hyperparameter Search Space}
\label{tab:tabr_hyperparams}
\begin{tabular}{lll}
\toprule
\textbf{Hyperparameter} & \textbf{Type} & \textbf{Range} \\
\midrule
\multicolumn{3}{l}{\textit{Model Parameters}} \\
d\_main & UniformInt & [96, 384] \\
context\_dropout & Uniform & [0.0, 0.6] \\
dropout0 & Uniform & [0.0, 0.6] \\
dropout1 & Fixed & 0.0 \\
encoder\_n\_blocks & Categorical & \{0, 1\} \\
predictor\_n\_blocks & Categorical & \{1, 2\} \\
\midrule
\multicolumn{3}{l}{\textit{Optimizer}} \\
optimizer type & Fixed & AdamW \\
learning rate & Log-uniform & [$3 \times 10^{-5}$, $10^{-3}$] \\
weight decay & Categorical+Log-uniform & \{0, [$10^{-6}$, $10^{-4}$]\} \\
\midrule
\multicolumn{3}{l}{\textit{Embeddings}} \\
embedding type & Fixed & PLREmbeddings \\
n\_frequencies & UniformInt & [16, 96] \\
d\_embedding & UniformInt & [16, 64] \\
frequency\_scale & Log-uniform & [$10^{-2}$, $10^{2}$] \\
lite & Fixed & True \\
\bottomrule
\end{tabular}
\end{table*}

\paragraph{RealMLP}

The hyperparameter search space for RealMLP \cite{holzmuller2024better} on the TabZilla Hard Benchmark \cite{mcelfresh2023tabzilla} is presented in \cref{tab:realmlp_hyperparams}. These hyperparameter ranges were selected based on the recommendations provided in the official RealMLP repository on GitHub\footnote{https://github.com/dholzmueller/pytabkit}.

\begin{table*}[htbp]
\centering
\caption{RealMLP Hyperparameter Search Space}
\label{tab:realmlp_hyperparams}
\begin{tabular}{lll}
\toprule
\textbf{Hyperparameter} & \textbf{Type} & \textbf{Range} \\
\midrule
\multicolumn{3}{l}{\textit{Model Architecture}} \\
num\_emb\_type & Categorical & \{none, pbld, pl, plr\} \\
add\_front\_scale & Categorical & \{True (0.6), False (0.4)\} \\
hidden\_sizes & Categorical & \{[256, 256, 256] (0.6), [64, 64, 64, 64, 64] (0.2), [512] (0.2)\} \\
act & Categorical & \{relu, selu, mish\} \\
\midrule
\multicolumn{3}{l}{\textit{Training Parameters}} \\
lr & Log-uniform & [$2 \times 10^{-2}$, $3 \times 10^{-1}$] \\
p\_drop & Categorical & \{0.0 (0.3), 0.15 (0.5), 0.3 (0.2)\} \\
wd & Categorical & \{0.0, $2 \times 10^{-2}$\} \\
ls\_eps & Categorical & \{0.0 (0.3), 0.1 (0.7)\} \\
\midrule
\multicolumn{3}{l}{\textit{PLR Embedding}} \\
plr\_sigma & Log-uniform & [0.05, 0.5] \\
\bottomrule
\end{tabular}
\end{table*}

\paragraph{TabM}

The hyperparameter search space for TabM \cite{gorishniy2025tabm} on the TabZilla Hard Benchmark \cite{mcelfresh2023tabzilla} is presented in \cref{tab:tabm_hyperparams}. These hyperparameter ranges were selected based on the recommendations provided in the paper.

\begin{table*}[htbp]
\centering
\caption{TabM Hyperparameter Search Space}
\label{tab:tabm_hyperparams}
\begin{tabular}{lll}
\toprule
\textbf{Hyperparameter} & \textbf{Type} & \textbf{Range} \\
\midrule
\multicolumn{3}{l}{\textit{General Parameters (Both Cases)}} \\
num\_emb\_type & Categorical & \{none, pwl\} \\
arch\_type & Categorical & \{tabm, tabm-mini\} \\
k & Fixed & 32 \\
d\_block & UniformInt & [64, 1024] \\
dropout & Categorical+Uniform & \{0, [0.1, 0.5]\} \\
weight\_decay & Categorical+Log-uniform & \{0, [$10^{-4}$, $10^{-1}$]\} \\
\midrule
\multicolumn{3}{l}{\textit{When num\_emb\_type = none}} \\
n\_blocks & UniformInt & [1, 6] \\
lr & Log-uniform & [$10^{-4}$, $5 \times 10^{-3}$] \\
\midrule
\multicolumn{3}{l}{\textit{When num\_emb\_type = pwl}} \\
n\_blocks & UniformInt & [1, 5] \\
lr & Log-uniform & [$5 \times 10^{-5}$, $3 \times 10^{-3}$] \\
num\_emb\_n\_bins & UniformInt & [8, 32] \\
\bottomrule
\end{tabular}
\end{table*}

\paragraph{ModernNCA}

The hyperparameter search space for ModernNCA \cite{ye2025modernnca} on the TabZilla Hard Benchmark \cite{mcelfresh2023tabzilla} is presented in \cref{tab:modernnca_hyperparams}. These hyperparameter ranges were selected based on the recommendations provided in the official ModernNCA repository on GitHub\footnote{https://github.com/LAMDA-Tabular/TALENT}.

\begin{table*}[htbp]
\centering
\caption{ModernNCA Hyperparameter Search Space}
\label{tab:modernnca_hyperparams}
\begin{tabular}{lll}
\toprule
\textbf{Hyperparameter} & \textbf{Type} & \textbf{Range} \\
\midrule
\multicolumn{3}{l}{\textit{Model Architecture}} \\
dim & UniformInt & [64, 1024] \\
dropout & Uniform & [0.0, 0.5] \\
d\_block & UniformInt & [64, 1024] \\
n\_blocks & Categorical & \{0, 1, 2\} \\
temperature & Fixed & 1 \\
sample\_rate & Uniform & [0.05, 0.6] \\
\midrule
\multicolumn{3}{l}{\textit{Training Parameters}} \\
lr & Log-uniform & [$10^{-5}$, $10^{-1}$] \\
weight\_decay & Categorical+Log-uniform & \{0, [$10^{-6}$, $10^{-3}$]\} \\
\midrule
\multicolumn{3}{l}{\textit{PLR Embeddings}} \\
embedding\_type & Fixed & PLREmbeddings \\
n\_frequencies & UniformInt & [16, 96] \\
d\_embedding & UniformInt & [16, 64] \\
frequency\_scale & Log-uniform & [0.005, 10.0] \\
lite & Fixed & True \\
\bottomrule
\end{tabular}
\end{table*}

\paragraph{TabPFNv2}

The hyperparameter search space for TabPFNv2 \cite{hollmann2025accurate} on the TabZilla Hard Benchmark \cite{mcelfresh2023tabzilla} and the regression tasks is presented in \cref{tab:TabPFNv2_params}. These hyperparameter ranges were selected based on the default output of \texttt{hpo.TabPFNSearchSpace}'s \texttt{get\_classifier\_space()} and \texttt{get\_regressor\_space()}, as recommended by the official documentation.

\begin{table*}[htbp]
\centering
\caption{TabPFNv2 Hyperparameter Search Space}
\label{tab:TabPFNv2_params}
\begin{tabular}{llll}
\toprule
\textbf{Hyperparameter} & \textbf{Type} & \textbf{Range} & \textbf{Default} \\
\midrule
Number of estimators & UniformInt & [1, 8] & 4 \\
Softmax Temperature & Categorical & \{0.75, 0.8, 0.85, 0.9, 0.95, 1\} & 0.9 \\
Average before Softmax & Categorical & \{True, False\} & False \\
\bottomrule
\end{tabular}
\end{table*}

\clearpage
\clearpage

\section{Discarding Criteria for Meta-Training Datasets}
\label{sec:discarding-criteria}

Meta-training a model on a large collection of diverse real tabular datasets requires careful dataset selection to avoid data leakage and ensure generalizability. 
In this section, we describe our rigorous criteria for discarding datasets (\ie excluding from our meta-training collection $\cM_{\text{train}}$) to ensure \emph{no overlap} and \emph{minimizing any form of implicit leakage} with evaluation datasets. Below, we detail all the steps and checks involved in the discarding process.

\subsection{Fundamental Constraints and Motivation}

To maintain the integrity of our experimental setup, we enforce the following fundamental constraints:
\begin{itemize}
    \item No dataset used for evaluation appears in the meta-training collection.
    \item Datasets that are highly similar to evaluation datasets, in name or structure, are discarded.
    \item Edge-case datasets that may introduce inconsistencies or unexpected biases are removed.
    \item Sample-level checks ensure no data leakage between meta-training and evaluation sets.
\end{itemize}

Given the complexity of tabular data hosting sites, where multiple versions and slight variations of the same dataset can exist, our discarding strategy extends beyond simple name filtering to include structural similarity and content-based exclusions. We address subtleties such as the presence of multiple copies of the same dataset (sometimes with slight differences in naming, features, or samples), and potential sample-level duplication.

\subsection{High-Level Overview}

We begin with a large unfiltered set $\cP$ of all open-source tabular classification datasets in OpenML \cite{OpenML2013, feurer2021openml, bischl2025openml}. Then:

\begin{enumerate}
    \item \textbf{Identification of evaluation datasets:} We define a set of \emph{evaluation datasets} $\cE$, as a reference panel for what must be excluded from the meta-training collection $\cM_{\text{train}}$ to avoid data leakage.

    \item \textbf{Metadata Loading:} We read metadata about each candidate dataset (\eg name, number of features, number of instances), together with its unique dataset identifier (\emph{did}).
    \item \textbf{Discard File Generation:} For each combination of ``metadata'' source and ``evaluation datasets'' we build a \emph{discard file} that records whether a dataset is excluded and the reason for exclusion.
    \item \textbf{Name-Based Discards:} We remove any dataset whose name (or partial/fuzzy variation) matches the name of an evaluation dataset. We also remove datasets whose name contains any evaluation dataset name as a substring, and any duplicates that differ only by version tags or small string variations. 
    
    \item \textbf{Size/Property-Based Discards:} Datasets with (i)~too few or too many samples, (ii)~too few or too many features, or (iii)~identical shape to known evaluation datasets are discarded. We do not take into account the number of classes, meaning that we discard any potential dataset that has the same number of samples and features as any of the evaluation datasets. Although this may discard too many datasets that could actually be used for meta-training, we decided to keep this strong rule to avoid unexpected leakage.
    \item \textbf{Sample-Level Checks:} Finally, we take a small sample (typically 5 random rows) from each evaluation dataset and compare it against each possible candidate dataset row to detect \emph{sample-level} leakage.
    \item \textbf{Final Inclusion Decision:} The remaining datasets, after all discards, form our meta-training collection $\cM_{\text{train}}$.
\end{enumerate}

\subsection{Multi-Level Discarding Pipeline}
The discarding pipeline consists of multiple stages, each imposing stricter criteria to filter out potential issues. 

\subsubsection{Evaluation Datasets Exclusion}
We first ensure that any dataset \emph{explicitly} used for evaluation is discarded. Let $\cE$ represent the set of evaluation datasets, as the union of the meta-validation datasets $\cM_{\text{val}}$ used for internal evaluation during meta-training, and the final meta-test collection $\cM_{\text{test}}$ used for final evaluation and benchmarking. These sets are disjoint:

\[
\cE = \cM_{\text{val}} \cup \cM_{\text{test}}, \quad \cM_{\text{val}} \cap \cM_{\text{test}} = \emptyset.
\]
To obtain an intermediate filtered set of potential meta-training datasets, we discard all datasets that appear in \(\cE\) from the initial unfiltered collection \(\cP\):

\begin{equation}
    \cM_{\text{train}}' = \cP \setminus \bigcup_{\cD \in \cE} \cD
\end{equation}
This first ensures that all explicitly known evaluation datasets are removed from meta-training.

\subsubsection{Dataset Name Similarity Checks}
Since OpenML contains multiple versions of the same dataset under slightly different names, we apply both exact and approximate matching techniques to identify and remove potential duplicates.

\paragraph{Exact Name Matching:} We define a function \(\sigma: \mathcal{S} \to \mathcal{S}\) that maps a dataset name to its sanitized version (\eg lowercased and normalized), where \(\mathcal{S}\) is the set of all possible dataset names. Let \(n_{\cD}\) denote the name of a dataset \(\cD\), and \(\mathcal{N}_{\cE}^\sigma\) be the set of sanitized names corresponding to the evaluation datasets in \(\cE\):

\begin{equation}
    \mathcal{N}_{\cE}^\sigma = \{\sigma(n_{\cD}) \mid \cD \in \cE\}.
\end{equation}

A dataset \(\cD \in \cM_{\text{train}}'\) is discarded if its sanitized name belongs to \(\mathcal{N}_{\cE}^\sigma\):

\begin{equation}
    \forall \cD \in \cM_{\text{train}}': \sigma(n_{\cD}) \in \mathcal{N}_{\cE}^\sigma, \implies \cM_{\text{train}}' \gets \cM_{\text{train}}' \setminus \{\cD\}.
\end{equation}

\paragraph{Cleaning Common Keywords:} To catch variations in dataset names, we define a preprocessing function \(\tau: \mathcal{S} \to \mathcal{S}\) that removes common keywords such as \texttt{small}, \texttt{ medium}, \texttt{ processed}, \texttt{ classif}, \texttt{ regression}, \texttt{ version}, etc., as well as years (\eg 2016, 2017), digits, and other non-informative patterns. This is achieved using regular expressions (regex) to systematically strip out numbers and other non-alphabetical elements. Let \(\mathcal{N}_{\cE}^{\tau}\) be the set of sanitized names of evaluation datasets after applying this transformation:

\begin{equation}
    \mathcal{N}_{\cE}^{\tau} = \{\tau(n_{\cD}) \mid \cD \in \cE\}.
\end{equation}

A dataset \(\cD \in \cM_{\text{train}}'\) is discarded if its transformed name belongs to \(\mathcal{N}_{\cE}^{\tau}\):

\begin{equation}
    \forall \cD \in \cM_{\text{train}}', : \tau(n_{\cD}) \in \mathcal{N}_{\cE}^{\tau}, \implies \cM_{\text{train}}' \gets \cM_{\text{train}}' \setminus \{\cD\}.
\end{equation}

\paragraph{Substring Matches:} We remove any dataset \(\cD \in \cM_{\text{train}}'\) whose name \(n_{\cD}\) contains the name of any evaluation dataset \(\cD' \in \cE\) as a substring. Formally, let \(n_{\cD'}\) denote the name of an evaluation dataset \(\cD' \in \cE\). A dataset \(\cD \in \cM_{\text{train}}'\) is discarded if there exists \(\cD' \in \cE\) such that \(n_{\cD'}\) is a substring of \(n_{\cD}\):

\begin{equation}
    \forall \cD \in \cM_{\text{train}}': \; \exists \cD' \in \cE \,|\, n_{\cD'} \subseteq n_{\cD}, \implies \cM_{\text{train}}' \gets \cM_{\text{train}}' \setminus \{\cD\}.
\end{equation}

\paragraph{Approximate String Matching Using Similarity Measures:} To account for minor variations in dataset names, we employ two similarity measures to identify and discard datasets with names that are approximately similar to those in $\cE$:
\begin{itemize}
    \item \textbf{Levenshtein Similarity:} Let \(L_{\text{sim}}: \mathcal{S} \times \mathcal{S} \to [0, 1]\) denote the normalized Levenshtein similarity, defined as:
    \begin{equation}
        L_{\text{sim}}(n_1, n_2) = 1 - \frac{L(n_1, n_2)}{\max(|n_1|, |n_2|)},
    \end{equation}
    where \(L(n_1, n_2)\) is the Levenshtein distance \cite{Levenshtein} between strings \(n_1\) and \(n_2\), which measures the minimum number of edits (insertions, deletions, or substitutions) required to transform one string into the other, and \(|n_1|, |n_2|\) are their respective lengths. A dataset \(\cD \in \cM_{\text{train}}'\) is discarded if its name \(n_{\cD}\) satisfies:
    \begin{equation}
        L_{\text{sim}}(n_{\cD}, n_{\cD'}) \geq \lambda, \quad \exists \cD' \in \cE,
    \end{equation}
    where \(\lambda \in [0, 1]\) is a similarity threshold.

    \item \textbf{Token Sort Ratio (TSR)\footnote{\url{https://github.com/seatgeek/thefuzz}}:} Let \(\text{TSR}: \mathcal{S} \times \mathcal{S} \to [0, 1]\) denote the token sort ratio, a fuzzy matching technique that tokenizes, reorders, and compares the words in two names in a slightly different manner than $L_{\text{sim}}$. A dataset \(\cD \in \cM_{\text{train}}'\) is discarded if its name \(n_{\cD}\) satisfies:
    \begin{equation}
        \text{TSR}(n_{\cD}, n_{\cD'}) \geq \lambda, \quad \exists \cD' \in \cE.
    \end{equation}
\end{itemize}

We use a similarity threshold of \(\lambda = 0.8\) for both measures. Datasets whose names exceed this threshold for any of the two similarity metrics, for any evaluation dataset \(\cD' \in \cE\) are discarded, as they indicate high redundancy with evaluation datasets. This approach ensures that even slightly renamed or partially obfuscated duplicates are recognized.

\subsubsection{Structural Filtering}
Datasets are further discarded if their structural properties are too similar to evaluation datasets. Let \(N_{\cD}\) and \(F_{\cD}\) denote the number of instances and features, respectively, of a dataset \(\cD\). 
Given an evaluation dataset \(\cD' \in \cE\) with \(N_{\cD'}\) instances and \(F_{\cD'}\) features, a candidate training dataset \(\cD \in \cM_{\text{train}}'\) is discarded if it shares the same number of samples and features as \(\cD'\):

\begin{equation}
    \left( N_{\cD} = N_{\cD'} \right) \land \left( F_{\cD} = F_{\cD'} \right), \quad \exists \cD' \in \cE.
\end{equation}
This prevents subtle leakage from datasets that might be completely relabeled but structurally identical. Although this approach can be overly conservative in certain edge cases (\eg distinct datasets that happen to have identical shapes), we enforce it to avoid potential near-duplicates or disguised versions of the same data.

\subsubsection{Edge Case Removal}

Independently of potential overlap with evaluation datasets, we remove certain outliers from meta-training based on their structural properties. A dataset \(\cD \in \cM_{\text{train}}'\) is discarded if it satisfies any of the following conditions:
\begin{itemize}
    \item Fewer than \(2\) features: $F_{\cD} < 2$.
    \item Fewer than \(10\) samples: $N_{\cD} < 10$.
    \item More than \(100,\!000\) features: $F_{\cD} > 100,\!000$.
    \item More than \(1,\!000,\!000\) samples: $N_{\cD} > 1,\!000,\!000$.
\end{itemize}

\subsubsection{Sample-Level Leakage Detection}
One of the more stringent steps in our pipeline is a final row-by-row comparison of actual data samples between meta-training and evaluation datasets. This ensures that no training data is implicitly represented in evaluation benchmarks.

Let \(\cD' \in \cE\) denote an evaluation dataset with feature matrix \(\bX_{\cD'}\), where each row \(\bx_{\cD'} \in \bX_{\cD'}\) represents a sample. 
For each \(\cD'\), we sample \(k = 5\) rows uniformly at random to form a subset \(S_{\cD'} = \{\bx_{\cD'}^{(1)}, \dots, \bx_{\cD'}^{(k)}\}\). Let \(\bX_{\cD}\) be the feature matrix of a candidate training dataset \(\cD \in \cM_{\text{train}}'\), combining all its splits (train, validation, test). We discard $\cD$ if \textbf{any} of the sampled evaluation rows matches one of its rows:

\begin{equation}
    \exists \cD' \in \cE, \exists \bx_{\cD'} \in S_{\cD'}, \exists \bx_{\cD} \in \bX_{\cD} \,|\, \bx_{\cD'} \equiv \bx_{\cD} \land F_{\cD} = F_{\cD'},
\end{equation}
where \(\bx_{\cD'} \equiv \bx_{\cD}\) denotes that the multiset of feature values in \(\bx_{\cD'}\) is identical to that of \(\bx_{\cD}\), regardless of column order. We use a multiset (counter-based) comparison of feature values to detect matches irrespective of the original column ordering. Note that the candidate dataset \(\cD\) is discarded immediately upon detection of any matching row to save computation time.

Although this check can be computationally expensive to run, it provides an extra safeguard against including any datasets that merely differ in feature-column arrangement but contain identical samples.

\subsection{Conclusion}

The final meta-training set \(\cM_{\text{train}}\) consists of datasets in \(\cM_{\text{train}}'\) that pass all discarding conditions. Defining each discard criterion as \(\mathrm{DC}_i(\cD)\), where \(\mathrm{DC}_i(\cD)\) means dataset \(\cD\) meets the \(i\)-th condition, we have:
\begin{equation}
    \cM_{\text{train}} 
    \;=\; 
    \bigl\{\, \cD \in \cM_{\text{train}}' \;\bigm|\; \neg \bigl(\mathrm{DC}_1(\cD)\;\lor\;\mathrm{DC}_2(\cD)\;\lor\;\dots\bigr) \bigr\}.
    \label{eq:final-meta-training-set}
\end{equation}

This ensures that only datasets not meeting any of the discard criteria are included in the final meta-training set. By following this detailed pipeline, we achieve a robust filtering of meta-training data, completely avoiding duplicates or near-duplicates of evaluation datasets. The procedure we describe is conservative but necessary to prevent inadvertent data leakage in large-scale tabular meta-learning experiments.

\end{document}